\documentclass[final,3p,times,number]{elsarticle}

\usepackage{amsmath,amsfonts,amssymb}
\usepackage{algorithmic}
\usepackage{algorithm}
\usepackage{array}
\usepackage{textcomp}
\usepackage{url}
\usepackage{graphicx}
\usepackage[table]{xcolor}
\usepackage{subcaption}
\usepackage{hyperref}
\usepackage{multirow}
\usepackage{booktabs}
\usepackage{lineno}

\journal{Transportation Research Part C}

\begin{document}

\begin{frontmatter}

\title{CoInfra: A Large-Scale Cooperative Infrastructure Perception System and Dataset for Vehicle--Infrastructure Cooperation in Adverse Weather}

\author[uwaterloo]{Minghao~Ning}
\author[uwaterloo]{Yufeng~Yang}
\author[uwaterloo]{Keqi~Shu\corref{cor1}}
\cortext[cor1]{Corresponding author}
\author[uwaterloo]{Shucheng~Huang}
\author[uwaterloo]{Jiaming~Zhong}
\author[uwaterloo]{Maryam~Salehi}
\author[uwaterloo]{Mahdi~Rahmani}
\author[uwaterloo]{Jiaming~Guo}
\author[unb]{Yukun~Lu}
\author[hku]{Chen~Sun}
\author[rogers]{Aladdin~Saleh}
\author[ualberta]{Ehsan~Hashemi}
\author[uwaterloo]{Amir~Khajepour}

\affiliation[uwaterloo]{organization={Department of Mechanical and Mechatronics Engineering, University of Waterloo},
            city={Waterloo},
            state={Ontario},
            country={Canada}}
\affiliation[unb]{organization={Department of Mechanical Engineering, University of New Brunswick},
            city={Fredericton},
            state={New Brunswick},
            country={Canada}}
\affiliation[hku]{organization={Department of Data and Systems Engineering, University of Hong Kong},
            city={Hong Kong},
            state={Hong Kong},
            country={China}}
\affiliation[rogers]{organization={Technology Partnerships and Innovations, Rogers Communications, Canada Inc.},
            city={Toronto},
            state={Ontario},
            country={Canada}}
\affiliation[ualberta]{organization={Department of Mechanical Engineering, University of Alberta},
            city={Edmonton},
            state={Alberta},
            country={Canada}}

\begin{abstract}
Vehicle--infrastructure (V2I) cooperative perception can substantially extend the range, coverage, and robustness of autonomous driving systems beyond the limits of onboard-only sensing, particularly in occluded and adverse-weather environments. However, its practical value is still difficult to quantify because existing benchmarks do not adequately capture large-scale multi-node deployments, realistic communication conditions, and adverse-weather operation. This paper presents \textbf{CoInfra}, a deployable cooperative infrastructure perception platform comprising 14 roadside sensor nodes connected through a commercial 5G network, together with a large-scale dataset and an open-source system stack for V2I cooperation research. The system supports synchronized multi-node sensing and delay-aware fusion under real 5G communication constraints. The released dataset covers an eight-node urban roundabout under four weather conditions (sunny, rainy, heavy snow, and freezing rain) and contains 294k LiDAR frames, 589k camera images, and 332k globally consistent 3D bounding boxes. It also includes a synchronized V2I subset collected with an autonomous vehicle. 
Beyond standard perception benchmarks, we further evaluate whether infrastructure sensing improves awareness of safety-critical traffic participants during roundabout interactions. In structured conflict scenarios, V2I cooperation increases critical-frame completeness from 33\%--46\% with vehicle-only sensing to 86\%--100\%. These results show that multi-node infrastructure perception can significantly improve situational awareness in conflict-rich traffic scenarios where vehicle-only sensing is most limited.
\end{abstract}


\end{frontmatter}

\section{Introduction}
\label{sec:introduction}

\begin{figure*}[t]
  \centering
  \includegraphics[width=0.95\textwidth]{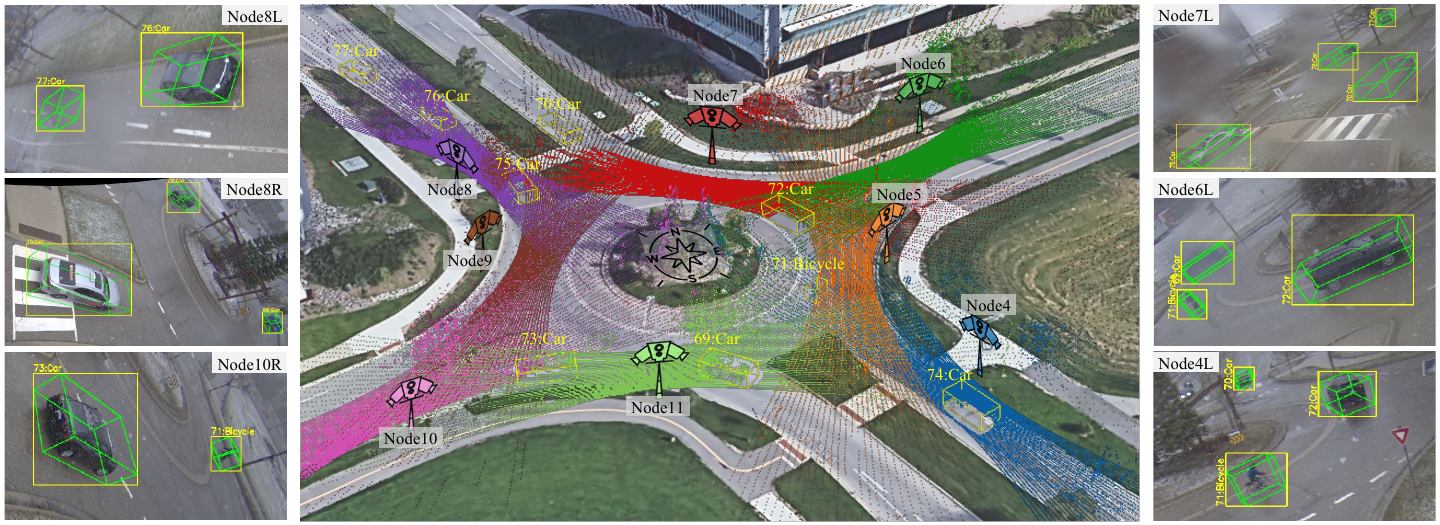}
  \caption{
    \textbf{Multi-node, multi-modal sample from the CoInfra dataset during heavy snow.}
    The central panel shows the global scene observed by eight infrastructure sensor nodes around a roundabout. Different colors represent LiDAR point clouds from different nodes. Objects are annotated with globally consistent 3D bounding boxes, IDs, and classes. Side panels show camera views from multiple nodes with projected 3D boxes. Adverse weather introduces LiDAR noise and camera blur, while occlusions and limited FOV challenge single-node perception. Cooperative fusion across nodes enables robust, globally consistent object detection and tracking under diverse and challenging conditions.
  }
  \label{fig:dataset_example}
\end{figure*}

Reliable perception remains a central bottleneck for autonomous driving in complex urban environments. Although recent advances in perception, prediction, and planning have substantially improved vehicle autonomy~\citep{ning2024novel, bhatt2024watonobus, ning2024efficient}, robust scene understanding is still difficult when traffic participants are partially occluded, distributed over a large area, or observed under adverse weather. These challenges are particularly significant at interaction-rich locations such as roundabouts and pedestrian crossings, where safe decision-making depends on timely awareness of all nearby conflict participants.

Vehicle--infrastructure cooperation offers a promising way to overcome these viewpoint limitations by augmenting onboard sensing with elevated roadside observations~\citep{yang2025real, xiang2023multi, huang2022multi}. In such systems, infrastructure sensor nodes perform local perception and transmit their outputs to a central cloud, which aggregates observations across nodes and broadcasts the fused scene representation to connected vehicles. 
The potential is substantial: infrastructure sensors have elevated, complementary viewpoints, allowing them to detect road users that are occluded or beyond the sensing range of an individual vehicle.
However, realizing this potential in practice requires studying cooperative perception not in isolation, but as a deployable system subject to real communication delays, adverse weather, and the interaction dynamics of real traffic.

Despite rapid progress in cooperative perception algorithms, three gaps continue to limit real-world research and deployment. First, synchronization across distributed nodes whose outputs arrive with stochastic delays is often treated as an implementation detail rather than an explicit object of study~\citep{xiang2024v2xreal, zimmer2024tumtraf, axmann2023lucoop}. Second, reproducible full-stack systems remain scarce: most prior work releases either datasets or algorithmic results, but not the integrated sensing, calibration, synchronization, communication, and software tools needed to replicate or extend large-scale deployments~\citep{yu2022dair, xu2023v2v4real, li2022v2x, xiang2024v2xreal}. Third, existing cooperative perception datasets remain limited in scale, agent count, and environmental diversity, with adverse weather still underrepresented in real-world benchmarks~\citep{Karvat_2024_AdverCity}. 
Critically, even when these technical barriers are addressed, evaluation in prior work has been dominated by detector-level metrics such as average precision, with much less attention to whether cooperative infrastructure sensing improves the situational awareness needed for safe driving at conflict-rich locations.

This paper presents \textbf{CoInfra}, a large-scale cooperative infrastructure perception platform that addresses these gaps. The platform comprises 14 roadside sensor nodes connected through a commercial 5G network, each equipped with LiDAR, cameras, and edge computing. 
The deployment is centered on an urban roundabout, chosen because it creates frequent merging, yielding, line-of-sight conflict, and vulnerable-road-user (VRU) interactions. On top of this deployment, we construct a multi-node dataset with synchronized LiDAR and camera data collected under sunny, rainy, heavy snow, and freezing rain conditions, together with globally consistent 3D annotations and an aligned HD map.

The central claim of this work is that \emph{real-world vehicle–infrastructure cooperation should be studied as a deployable sensing system under communication delay and adverse weather, and its value should be assessed through safety-relevant observability in conflict-rich traffic interactions, not only detector AP.} Our contributions are:

\begin{enumerate}
  \item \textbf{Deployable cooperative infrastructure platform and dataset.} We develop and release an open-source cooperative infrastructure perception system consisting of 14 roadside sensor nodes with commercial 5G connectivity, together with a large-scale dataset containing 294k LiDAR frames, 589k camera images, and 332k globally consistent 3D bounding boxes collected from eight nodes under four weather conditions. We also release the full system stack, including hardware documentation, calibration and synchronization tools, management UI, and an Autoware\cite{autoware}-compatible ROS perception pipeline.

  \item \textbf{Delay-aware synchronization under realistic communication constraints.} We formulate multi-node synchronization as a probabilistic completeness--latency trade-off under stochastic 5G delay, and develop an adaptive fusion-window policy. Real-world communication measurements characterize the operational envelope, and trace-driven experiments show that adaptive synchronization with lightweight state prediction reduces tail latency while preserving fusion completeness.

  \item \textbf{Transportation-facing evaluation through ego-centric V2I observability.} Beyond standard perception benchmarks, we introduce an ego-centric observability analysis that evaluates whether infrastructure sensing restores awareness of safety-critical participants in structured roundabout conflict scenarios. In the most demanding cases, vehicle-only critical-frame completeness is as low as 33\%--46\%, while combined V2I sensing raises it to 86\%--100\%. This result demonstrates that the platform's value extends well beyond detector-level AP to the situational awareness on which safe driving decisions depend.
\end{enumerate}






The remainder of this paper is organized as follows. Section~\ref{sec:related_work} reviews related work and positions CoInfra relative to prior datasets and systems. Section~\ref{sec:problem_formulation} formulates the synchronization problem. Section~\ref{sec:system_design} describes the deployed platform and its open-source release. Section~\ref{sec:dataset} presents the dataset. Section~\ref{sec:experiments} reports the experimental evaluation, culminating in the ego-centric observability analysis that demonstrates the platform's transportation-facing value. Section~\ref{sec:conclusion} concludes.

\section{Related Work}
\label{sec:related_work}

\subsection{Cooperative Perception Systems and Fusion Architectures}

Cooperative perception extends situational awareness by aggregating observations from multiple vehicles and/or infrastructure nodes~\citep{arnold2022cooperative, wang2020v2vnet, xu2022opv2v}. Existing systems generally fall into three architectural categories: vehicle-to-vehicle (V2V), vehicle-to-infrastructure (V2I), and hybrid settings combining both~\citep{huang2022multi, arnold2022cooperative}. Across these settings, the core design problem is how to distribute sensing, local processing, communication, and fusion while respecting bandwidth and latency constraints.

Related research has also explored fusion strategies. Early fusion aggregates raw sensor data before detection, maximizing geometric completeness but imposing heavy bandwidth and synchronization requirements~\citep{chen2019fcooper}. Intermediate fusion exchanges learned features or BEV representations to balance accuracy and communication efficiency~\citep{wang2020v2vnet, xu2022v2xvit, li2023disconet, xu2023cobevt, hu2022where2comm}. Late fusion shares high-level detections or tracks, reducing communication load at the cost of potentially weaker recovery of occluded or sparse objects~\citep{arnold2022cooperative}. This accuracy--bandwidth trade-off is central to deployable cooperative perception and motivates the benchmarking setup used in this paper.

At the system level, however, relatively few works address the practical challenges of real-world deployment. DAIR-V2X~\citep{yu2022dair}, LUCOOP~\citep{axmann2023lucoop}, TUMTraf V2X~\citep{zimmer2024tumtraf}, and V2X-Real~\citep{xiang2024v2xreal} represent important steps toward real-world cooperative sensing, but they differ substantially in scale, openness, and operational focus. In particular, synchronization and communication are often treated as implementation details rather than as explicit objects of study, and few systems characterize delay behavior under realistic wireless conditions.

\subsection{Communication-Aware Cooperative Perception Algorithms}

A large body of work has focused on communication-efficient cooperative perception. V2VNet~\citep{wang2020v2vnet}, When2com~\citep{liu2022when2com}, and Who2com~\citep{liu2020who2com} study selective communication policies that determine when and what to transmit. Subsequent methods, including V2X-ViT~\citep{xu2022v2xvit}, DiscoNet~\citep{li2023disconet}, CoBEVT~\citep{xu2023cobevt}, and Where2comm~\citep{hu2022where2comm}, improve feature fusion through attention, transformers, and spatial confidence mechanisms. Recent work has also considered heterogeneity across agents, as in HEAL~\citep{lu2024heal}, and multi-modal cooperative fusion, as in BM2CP~\citep{zhang2024bm2cp}.

Most of these methods are evaluated in simulation or under idealized synchronization assumptions. As a result, the communication channel is often modeled as bandwidth-limited but temporally well-behaved. In contrast, real deployments must cope with stochastic latency, variable link quality, and occasional delayed or missing messages. This gap between algorithmic assumptions and systems reality motivates the delay-aware fusion adopted in CoInfra.

\subsection{Datasets for Cooperative Perception}

\begin{table*}[t]
  \centering
  \caption{Summary of major cooperative perception datasets. 
    \textbf{Type}: Sim = simulation; Real = real-world. 
    \textbf{Agents}: V2V = vehicle-to-vehicle; V2I = vehicle-to-infrastructure; 
    I2I = infrastructure-to-infrastructure. 
    \textbf{Weather}: C = clear; R = rainy; F = freezing rain; S = snowy. 
    In the \textbf{\#3D Boxes} column, $^*$ denotes the total number of annotated 3D instances as reported by the original dataset, which may include repeated observations of the same object across agents, while $^\dagger$ denotes the number of globally unique boxes in a shared coordinate frame. 
    CoInfra consists of two subsets: \textbf{CoInfra-I2I}, the primary multi-node infrastructure dataset used for benchmarking, and \textbf{CoInfra-V2I}, a smaller synchronized infrastructure--vehicle subset used for vehicle--infrastructure cooperation experiments. The additional class for \textbf{CoInfra-I2I} is the ego autonomous shuttle.}
  \label{tab:cp_datasets}
  \scriptsize
  \begin{tabular}{l c c c c c c c}
    \toprule
    \textbf{Dataset} & \textbf{Type} & \textbf{\#Agents} & \textbf{\#LiDAR} & \textbf{\#Images} & \textbf{\#3D Boxes} & \textbf{\#Classes} & \textbf{Weather} \\
    \midrule
    OPV2V~\citep{xu2022opv2v} & Sim & 2--7 (V2V) & 11k & 44k & 232k$^*$ & 1 & C \\
    V2X-Sim~\citep{li2022v2x} & Sim & 2--6 (V2V+V2I) & 10k & 60k & 26k$^*$ & 1 & C \\
    AdverseCity~\citep{Karvat_2024_AdverCity} & Sim & 5 (V2V+V2I) & 120k & 480k & 890k$^*$ & 6 & CR \\
    \midrule
    DAIR-V2X~\citep{yu2022dair} & Real & 2 (V2I) & 39k & 39k & 464k$^*$ & 10 & CR \\
    V2V4Real~\citep{xu2023v2v4real} & Real & 2 (V2V) & 20k & 40k & 240k$^*$ & 5 & C \\
    TUMTraf V2X~\citep{zimmer2024tumtraf} & Real & 6 (V2V+V2I) & 2k & 5k & 30k$^*$ & 8 & C \\
    LUCOOP~\citep{axmann2023lucoop} & Real & 3 (V2V) & 54k & 0 & 90k$^*$ & 4 & C \\
    V2X-Real~\citep{xiang2024v2xreal} & Real & 4 (V2V+V2I) & 33k & 171k & 1.2M$^*$ & 10 & C \\
    V2X-Radar~\citep{yang2024v2x} & Real & 2 (V2I) & 20k & 40k & 350k$^*$ & 5 & CRS \\
    Rope3D~\citep{ye2022rope3d} & Real & 1 (I) & 50k & 50k & 1.5M$^*$ & 12 & CR \\
    IPS300+~\citep{wang2023ips300} & Real & 2 (I2I) & 28k & 57k & 200k$^*$ & 7 & C \\
    RCooper~\citep{hao2024rcooper} & Real & 2--4 (I2I) & 30k & 50k & -- & 10 & CR \\
    \midrule
    \textbf{CoInfra-I2I (Ours)} & Real & \textbf{8 (I2I)} & \textbf{294k} & \textbf{589k} & \textbf{332k}$^\dagger$ & 5 & \textbf{CRFS} \\
    \textbf{CoInfra-V2I (subset)} & Real & 8 (I2I) + 1 AV & 46k & 92k & 43k$^\dagger$ & 6 & C \\
    \bottomrule
  \end{tabular}
\end{table*}

Cooperative perception datasets can be broadly divided into simulation-based and real-world benchmarks. Simulation datasets such as OPV2V~\citep{xu2022opv2v}, V2X-Sim~\citep{li2022v2x}, and AdverseCity~\citep{Karvat_2024_AdverCity} support controlled evaluation of fusion and communication strategies, but they do not fully capture real sensor noise, calibration drift, communication variability, or deployment complexity. They are therefore valuable for algorithm development, but less informative for assessing deployability.

Real-world datasets offer stronger realism but remain limited in different ways. DAIR-V2X~\citep{yu2022dair} and V2V4Real~\citep{xu2023v2v4real} established important V2I and V2V benchmarks, respectively, yet both involve only a small number of cooperating agents. TUMTraf V2X~\citep{zimmer2024tumtraf} and V2X-Real~\citep{xiang2024v2xreal} expand agent count, but still provide limited weather diversity. Infrastructure datasets such as Rope3D~\citep{ye2022rope3d} and IPS300+~\citep{wang2023ips300} offer large-scale roadside perception, but not multi-node cooperative fusion. Table~\ref{tab:cp_datasets} summarizes these differences.

Real-world datasets offer stronger realism but remain limited in different ways. 
DAIR-V2X~\citep{yu2022dair} and V2V4Real~\citep{xu2023v2v4real} established important V2I and V2V benchmarks, respectively, yet both involve only a small number of cooperating agents. 
TUMTraf V2X~\citep{zimmer2024tumtraf} and V2X-Real~\citep{xiang2024v2xreal} expand agent count, but still provide limited weather diversity. 
Roadside perception datasets such as Rope3D~\citep{ye2022rope3d}, IPS300+~\citep{wang2023ips300}, and RCooper~\citep{hao2024rcooper}, broaden the study of infrastructure sensing, but they do not provide a large-scale multi-node cooperative benchmark with the adverse-weather diversity and shared-scene coverage. 
Table~\ref{tab:cp_datasets} summarizes these differences.

Taken together, prior datasets reveal three persistent gaps: limited real-world multi-node scale, limited adverse-weather coverage, and limited support for studying how communication delay interacts with fusion performance in deployed cooperative perception systems. CoInfra is designed to address these gaps through a unified platform that combines an eight-node real-world cooperative perception dataset, explicit delay-aware fusion analysis, adverse-weather coverage, and end-to-end vehicle integration.




\section{Problem Formulation}
\label{sec:problem_formulation}

We consider a cooperative infrastructure perception system consisting of $N$ fixed sensor nodes
$\{S_1,S_2,\ldots,S_N\}$ deployed at elevated roadside locations.
Each node $S_i$ is equipped with a LiDAR and $K$ RGB cameras and performs local perception at a fixed frequency $f$ (e.g., 10~Hz), producing structured outputs such as object detections and tracks.
These outputs are transmitted to a cloud server through a wireless network, where data from multiple nodes are fused to construct a global scene representation.

\subsection{Multi-Node Cooperative Perception}

Let $\mathbf{o}_i^{t_k}$ denote the perception output produced by node $S_i$ for the globally aligned time anchor $t_k = k/f$, where $k \in \mathbb{Z}^+$.
At the cloud server, the global perception result at time $t_k$ is obtained by fusing the outputs available when fusion is triggered:

\begin{equation}
\mathbf{G}^{t_k} =
\mathcal{F}\!\left(\left\{\mathbf{o}_i^{t_k} \mid i \in \mathcal{A}^{t_k}\right\},\; \mathbf{M}\right),
\label{eq:global_fusion}
\end{equation}

where $\mathcal{A}^{t_k} \subseteq \{1,\ldots,N\}$ is the set of nodes whose data has arrived by the fusion time,
$\mathbf{M}$ denotes the HD map providing spatial context, and $\mathcal{F}$ is the fusion operator.
Depending on system design, $\mathcal{F}$ may correspond to early fusion (raw sensor data), intermediate fusion (feature-level fusion), or late fusion (object-level aggregation), each with different communication and accuracy trade-offs.

\subsection{Synchronization Under Communication Delay}
\label{sec:delay_formulation}

In a distributed system, node outputs do not arrive at the cloud simultaneously.
Instead, each message experiences a delay $\delta_i^{t_k}$ consisting of local processing time and communication latency. As a result, the cloud server must decide how long to wait for node messages before triggering fusion.

\begin{equation}
\delta_i^{t_k} = \delta_i^{\text{proc}} + \delta_i^{\text{comm}}.
\label{eq:delay_decomp}
\end{equation}

To characterize this synchronization problem, we maintain an empirical latency model for each node using a sliding window of recent observations.
In practice, the observed delay distributions are reasonably approximated by a Gaussian model,

\begin{equation}
\delta_i \sim \mathcal{N}(\hat{\mu}_i,\hat{\sigma}_i^2),
\label{eq:latency_model}
\end{equation}
where $\hat{\mu}_i$ and $\hat{\sigma}_i$ are the online estimates of the mean and standard deviation.
This model is a practical approximation used to guide system design rather than a strict assumption required by the framework.

To synchronize node outputs, the cloud server introduces a fusion window $\Delta T$.
For each time anchor $t_k$, the server waits until time $t_k+\Delta T$ before triggering fusion.
Messages arriving within this window are included in $\mathcal{A}^{t_k}$.
Increasing $\Delta T$ improves the probability that more node messages are available for fusion.
Under the latency model in Eq.~(\ref{eq:latency_model}), the probability that all node messages arrive before the fusion trigger is shown as follows, where $\Phi(\cdot)$ is the standard normal cumulative distribution function.

\begin{equation}
P_{\text{complete}}(\Delta T)
=
\prod_{i=1}^{N}
P(\delta_i^{t_k}\le \Delta T)
=
\prod_{i=1}^{N}
\Phi\!\left(
\frac{\Delta T-\hat{\mu}_i}{\hat{\sigma}_i}
\right),
\label{eq:completeness}
\end{equation}


A larger fusion window, however, also increases reaction latency because the global perception result is released later.
This introduces a fundamental trade-off between \emph{fusion completeness} and \emph{system responsiveness}. In the deployed system, the fusion window is parameterized as:





\begin{equation}
\Delta T
=
\max_i\!\left(\hat{\mu}_i + n\hat{\sigma}_i\right),
\label{eq:fusion_window}
\end{equation}
where $n$ controls the confidence level that node messages will arrive within the window.
Based on empirical latency measurements, we use this formulation to guide the design of an adaptive fusion-window policy and use $n=2$ or $n=3$ as a conservative deployment setting, while Section~\ref{sec:sync_experiments} compares the performance under the measured latency regime.

Messages arriving after the main fusion trigger are not discarded.
Instead, they are incorporated through a post-fusion update.
Given an object state at time $t_k$, its current position is predicted as

\begin{equation}
\hat{\mathbf{p}}_i(t_{\text{current}})
=
\mathbf{p}_i(t_k)
+
\mathbf{v}_i\,(t_{\text{current}}-t_k),
\label{eq:motion_correction}
\end{equation}
where $\mathbf{v}_i$ is the estimated object velocity from the tracking module.
Late observations are integrated with reduced confidence to preserve temporal consistency without delaying the primary fusion output.

\section{System Design}
\label{sec:system_design}

\begin{figure*}[t]
  \centering
  \includegraphics[width=0.95\textwidth]{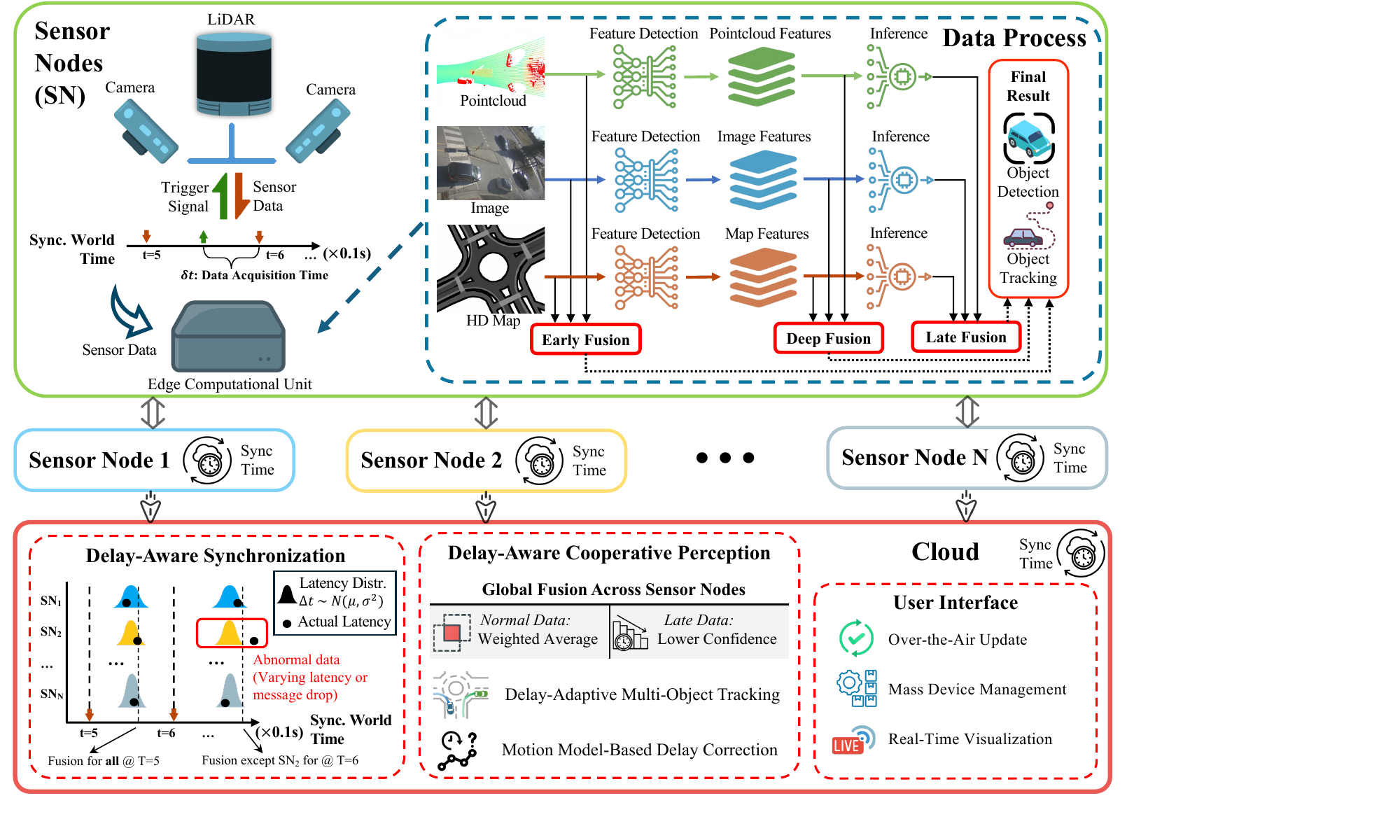}
  \caption{System overview and delay-aware synchronization in CoInfra. Each infrastructure sensor node acquires LiDAR and camera data at globally aligned time anchors, performs local perception at the edge, and transmits structured outputs to the cloud. The cloud estimates per-node latency online, applies an adaptive fusion window, and fuses node outputs into a global scene representation. Late-arriving data can be incorporated through post-fusion update.}
  \label{fig:time_sync}
\end{figure*}

This section describes the deployed CoInfra platform and its data production pipeline.
Fig.~\ref{fig:time_sync} summarizes the end-to-end architecture: each infrastructure node performs time-aligned sensing and local perception at the edge, transmits structured outputs through the 5G network, and contributes to cloud-level delay-aware fusion.

\subsection{Sensor Node Design}
\label{sec:sensor_node}

CoInfra comprises 14 infrastructure sensor nodes designed for continuous outdoor operation.
Each node integrates four key components: sensing, edge computing, communication, and power system.

\textbf{Sensing.}
Each node is equipped with one Robosense Helios 1615 LiDAR (32-beam, 200~m range, 10~Hz) and two Basler dart daA1920-160uc global-shutter RGB cameras (1920$\times$1200, up to 160~FPS, 79$^\circ$ horizontal field of view).

\textbf{Edge computing.}
Local perception is executed on an NVIDIA Jetson Orin NX (16~GB, 100~TOPS), which supports image-based detection, LiDAR clustering, multi-modal sensor fusion, and tracking in real time.

\textbf{Communication.}
Each node is connected to the cloud server through an industrial-grade 5G modem, enabling low-latency transmission of perception outputs as well as remote monitoring and software updates.

\textbf{Power system.}
Nodes are powered by solar panels, 12V battery, and charge controllers, enabling self-contained long-term deployment.

\begin{figure}[t]
    \centering
    \begin{minipage}[t]{0.48\columnwidth}
        \centering
        \includegraphics[width=0.8\linewidth]{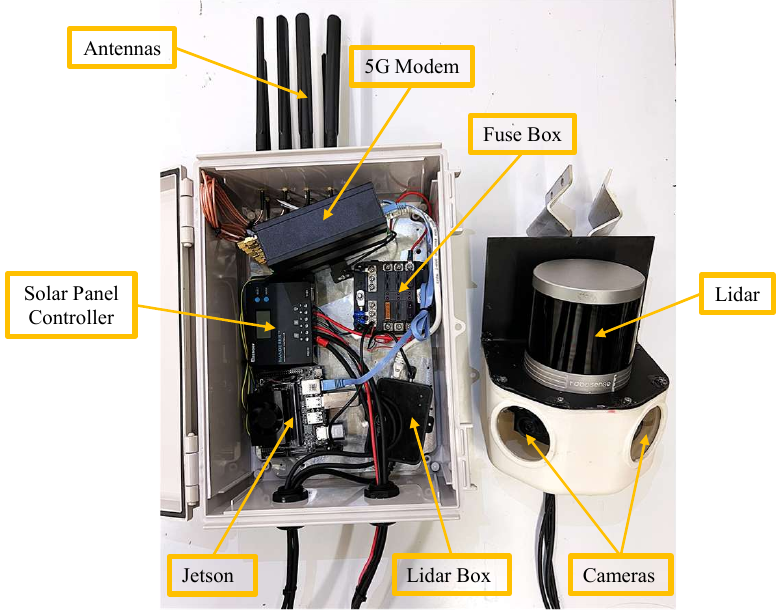}
        \caption{Structure of a single infrastructure sensor node, including LiDAR, dual RGB cameras, edge computing unit, and 5G modem.}
        \label{fig:one_node_material}
    \end{minipage}\hfill
    \begin{minipage}[t]{0.48\columnwidth}
        \centering
        \includegraphics[width=\linewidth]{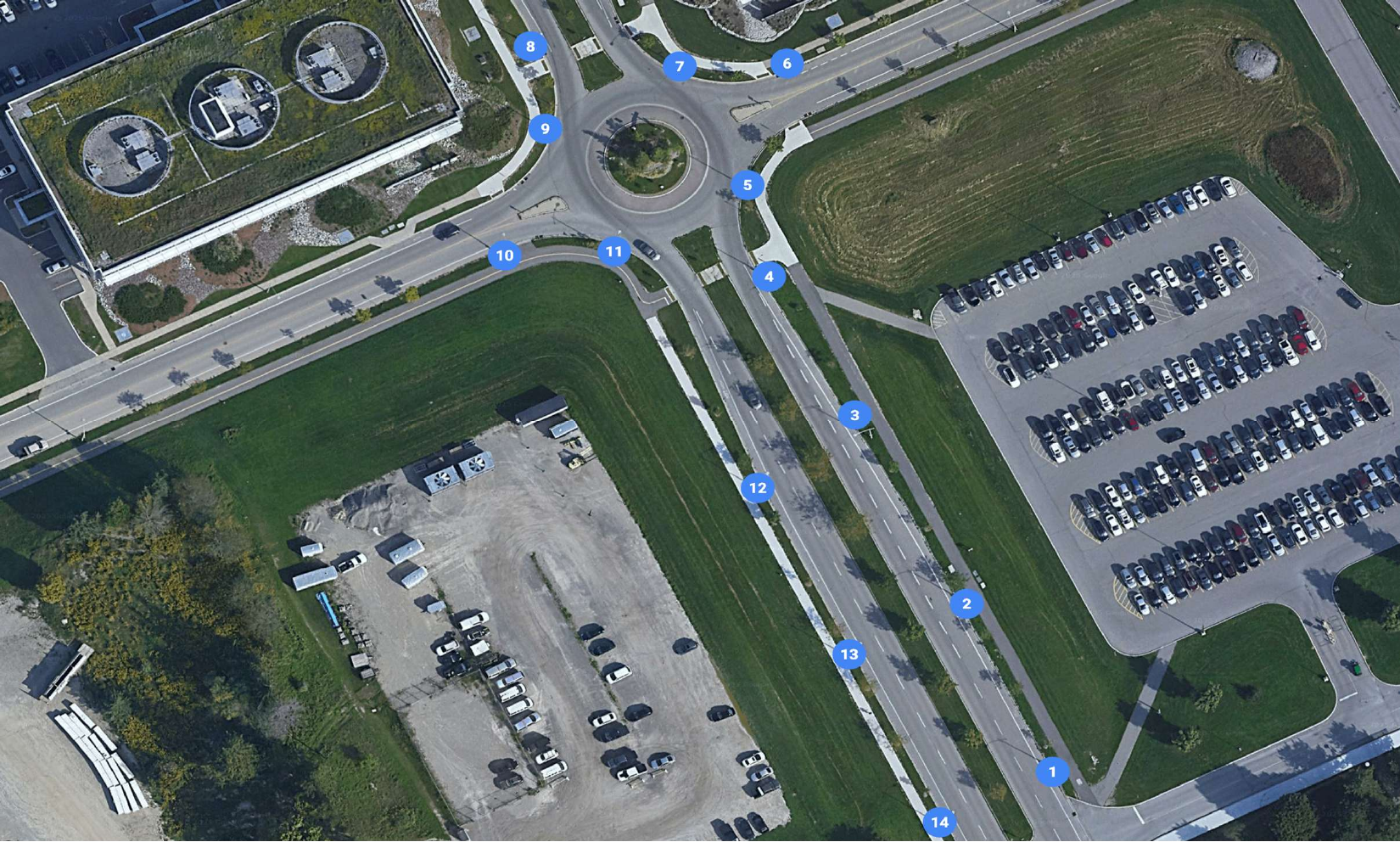}
        \caption{Deployment map of the 14 infrastructure sensor nodes on the University of Waterloo North Campus.}
        \label{fig:node_placement}
    \end{minipage}
\end{figure}

All components are mounted in weatherproof enclosures on a modular aluminum chassis (Fig.~\ref{fig:one_node_material}).

\subsection{Deployment}
\label{sec:deployment}

The 14 sensor nodes are installed on municipal light poles at approximately $7.5\,\mathrm{m}$ height across the University of Waterloo North Campus (Fig.~\ref{fig:node_placement}).
This elevated placement increases coverage, reduces foreground occlusion, and enables overlapping viewpoints across nearby nodes.
The deployment spans straight road segments, curved approaches, pedestrian crossings, sidewalks, and a multi-entry roundabout.
The roundabout is particularly suitable for cooperative perception research because it produces frequent merging, yielding, and VRU interactions while also creating line-of-sight limitations for single viewpoints.

\subsection{Calibration}
\label{sec:calibration}

Accurate multi-node fusion requires that all sensors be aligned within a shared global coordinate frame. 
To achieve this, we employ a multi-stage calibration pipeline that estimates both intra-node and global transformations.

\begin{enumerate}

\item \textbf{Camera intrinsic calibration.}  
Lens distortion parameters are estimated using a checkerboard-based calibration procedure~\citep{zhang2002flexible}.

\item \textbf{Camera--LiDAR extrinsic calibration.}  
The rigid transformation between each camera and LiDAR sensor is estimated using a checkerboard-based method that exploits 3D line and planar correspondences~\citep{zhou2018automatic}.

\item \textbf{Node-to-global initialization.}  
Each infrastructure node is roughly aligned to the global frame using AprilTag markers whose poses are measured by the autonomous shuttle’s high-precision localization system~\citep{bhatt2024watonobus}.

\item \textbf{Point cloud refinement.}  
The alignment is further refined by registering each node's LiDAR point cloud to the global HD map using the Iterative Closest Point (ICP) algorithm. 
The HD map itself is constructed from LiDAR data collected by the autonomous vehicle with high-precision localization.

\end{enumerate}

This calibration pipeline yields consistent geometric alignment across nodes, enabling reliable multi-node perception fusion. 
Calibration is repeated when necessary to account for possible sensor drift or mounting adjustments.

\begin{figure}[t]
\centering
\begin{minipage}[t]{0.48\columnwidth}
\centering
\includegraphics[width=\linewidth]{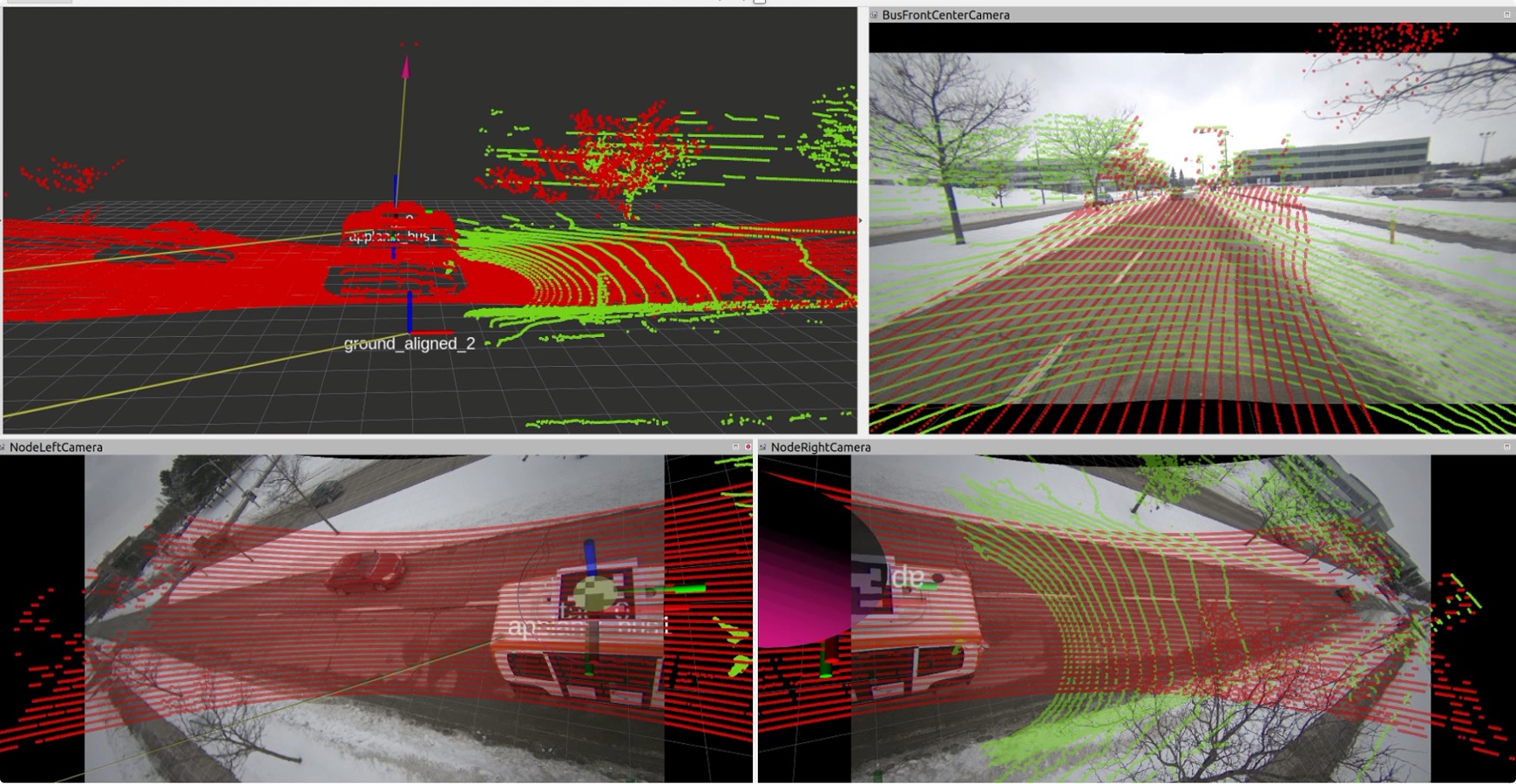}
\caption{Representative calibration results between a sensor node and the autonomous shuttle. 
Aligned LiDAR point clouds from the infrastructure node (red) and the shuttle (green) demonstrate consistent global registration. 
Camera views with projected LiDAR points further confirm accurate multi-modal alignment.}
\label{fig:calibration}
\end{minipage}\hfill
\begin{minipage}[t]{0.48\columnwidth}
\centering
\includegraphics[width=\linewidth]{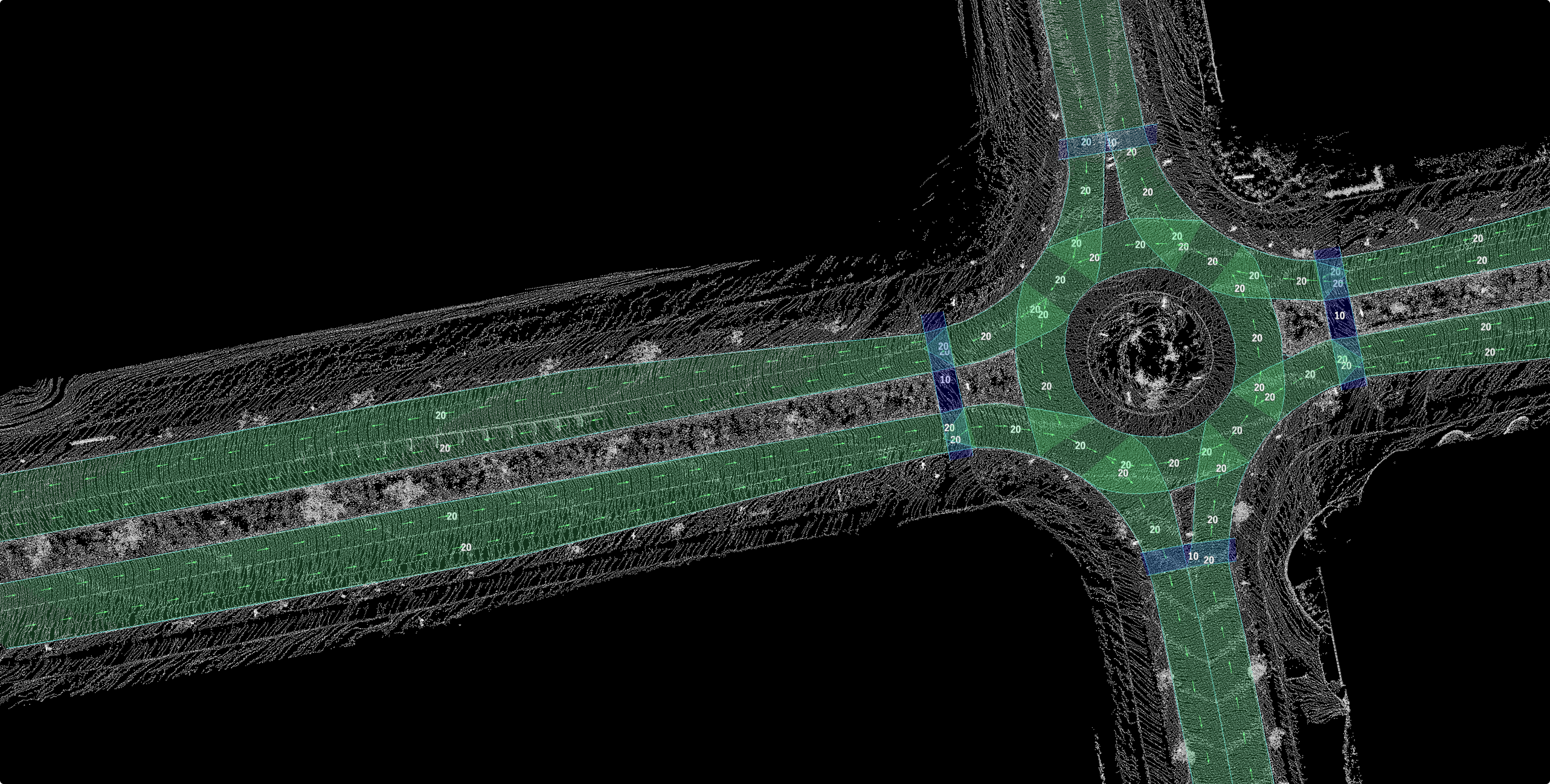}
\caption{High-definition map used as the global reference frame in the CoInfra deployment. 
The map is constructed from vehicle LiDAR data with high-precision localization, and serves as the spatial coordinate frame for aligning all infrastructure sensor observations.}
\label{fig:hdmap}
\end{minipage}
\end{figure}

Fig.~\ref{fig:calibration} illustrates representative alignment results between a sensor node and the autonomous shuttle, while Fig.~\ref{fig:hdmap} shows the HD map used as the global reference frame for all sensor data.

\subsection{Perception Pipeline}
\label{sec:perception_pipeline}

Each infrastructure node performs local perception at the edge and transmits structured outputs to the cloud rather than raw sensor streams. As shown in Section~\ref{sec:5g_analysis}, this design is necessary in practice: under multi-node operation, raw image transmission over 5G incurs multi-second latency and more than 60\% packet loss, whereas compact object-level outputs achieve sub-40~ms delivery with zero loss.

The deployed node-level pipeline uses LiDAR--camera detection with HD-map spatial priors. Specifically, LiDAR point clouds are projected into a bird's-eye-view (BEV) representation registered to the global HD map, providing both geometric input and spatial context for distinguishing drivable and non-drivable regions. The two camera views are also transformed and blended into an RGB BEV image. The resulting LiDAR and camera BEV channels are then concatenated and fed into a YOLOv11-OBB detector~\citep{tian2025yolov12}, followed by multi-object tracking~\citep{cao2023observation}. The resulting object detections and tracks are timestamped and transmitted to the cloud server, where delay-aware synchronization and multi-node fusion are performed as described in Section~\ref{sec:sync_protocol}. Because the platform is open-source, alternative fusion strategies, including intermediate feature-level fusion, can be integrated within the same system architecture.

\subsection{Delay-Aware Synchronization Protocol}
\label{sec:sync_protocol}

Following the formulation in Section~\ref{sec:delay_formulation}, the cloud server implements delay-aware synchronization through five operational steps.

\textbf{Clock alignment.}
All nodes are synchronized to the cloud server using NTP with \texttt{chrony}~\citep{dinar2020ntp}, providing a shared system clock for sensing and communication.

\textbf{Synchronized triggering.}
LiDAR and camera acquisition are scheduled at globally aligned time anchors spaced at 0.1-second intervals.
This reduces inter-node timestamp mismatch and ensures that local perception outputs correspond to approximately the same scene time.

\textbf{Local timestamping.}
Each node records the acquisition time of each frame using the shared system clock and attaches this timestamp to the corresponding structured perception output.

\textbf{Online latency estimation and fusion-window selection.}
At the cloud, recent message arrival times are used to estimate $\hat{\mu}_i$ and $\hat{\sigma}_i$ for each node.
These estimates determine the adaptive fusion window in Eq.~(\ref{eq:fusion_window}), which specifies how long the cloud waits before triggering fusion for each time anchor.

\textbf{Late-arrival handling.}
Messages arriving after the main fusion trigger may still be incorporated through post-fusion update using the motion-corrected state estimate in Eq.~(\ref{eq:motion_correction}), with reduced confidence.
This allows the system to preserve responsiveness while avoiding unnecessary loss of useful late observations.

\subsection{Open-Source Release}
\label{sec:open_source}

Unlike many prior cooperative perception datasets that release only logs and annotations, CoInfra also provides the deployable stack used to build and operate the system.
The release includes four major components:

\textbf{Node construction and calibration:} hardware documentation, assembly guidance, and software for intrinsic, extrinsic, and global-frame calibration;

\textbf{Synchronization and recording:} tools for clock alignment, synchronized sensing, distributed data recording, and cloud-side fusion-window management;

\textbf{System operation:} a web platform for fleet monitoring, remote management, OTA updates, and live visualization;

\textbf{ROS-based perception stack:} node- and cloud-level perception stack using Autoware-compatible message~\citep{autoware}.

The purpose of this release is not only to share a static dataset, but also to support replication, extension, and deployment of cooperative infrastructure perception systems in new environments.


\section{Dataset Description}
\label{sec:dataset}

\subsection{Data Collection}

For the dataset release, we selected eight sensor nodes that jointly provide full coverage of a complex urban roundabout environment containing vehicle lanes, pedestrian crossings, bicycle paths, and sidewalks. 
Data was collected continuously under diverse environmental conditions including sunny, rainy, heavy snow, and freezing rain. These conditions introduce realistic sensing challenges such as reduced visibility, LiDAR sparsity, and partial sensor occlusion.
Across the eight-node deployment, each acquisition cycle produces 16 RGB images and 8 LiDAR scans, corresponding to 160 images and 80 point clouds per second in total. From the collected recordings, representative clips were selected to capture a wide range of traffic scenarios, including vehicle interactions, VRU crossings, and roundabout merging behaviors.

\subsection{Dataset Subsets}

The CoInfra dataset is organized into two complementary subsets designed for different research purposes.
\textbf{CoInfra-I2I (Infrastructure-to-Infrastructure)} is the primary subset of the dataset and consists of synchronized observations from eight infrastructure sensor nodes. It captures multi-node perception under four weather conditions and contains the majority of the collected data.
Because the infrastructure network operates continuously and independently of any specific vehicle, this subset provides long-duration recordings of real-world traffic dynamics. It is therefore intended for large-scale benchmarking of cooperative perception tasks such as multi-node detection, tracking, and scene understanding.

\textbf{CoInfra-V2I (Vehicle-to-Infrastructure)} is a smaller subset collected during coordinated operation between the infrastructure sensor network and an Autoware-based autonomous shuttle bus~\citep{bhatt2024watonobus}. During these sessions, the shuttle traverses the roundabout while infrastructure nodes simultaneously record data.
This subset contains synchronized observations from both the infrastructure nodes and the vehicle’s onboard sensors, including LiDAR, cameras, and high-precision localization. It enables experiments on the complete vehicle–infrastructure cooperation loop, such as infrastructure-assisted perception and extended sensing range evaluation.
Because these coordinated sessions occur only when the vehicle enters the roundabout region, the V2I subset is smaller and is mainly intended for system-level validation rather than large-scale benchmarking.

\subsection{Data Annotation}

The dataset provides 3D bounding box annotations for five object classes: \textit{person}, \textit{bicycle}, \textit{car}, \textit{bus}, and \textit{truck}. All annotations are defined in a shared global coordinate frame, ensuring geometric consistency across all infrastructure nodes. The annotation pipeline consists of three main stages.
First, LiDAR point clouds from all nodes are transformed into the global frame. Ground points are removed and the remaining points are projected onto the bird’s-eye-view (BEV) plane.
Second, annotators label oriented 2D bounding boxes in the BEV representation using CVAT~\cite{cvat}. Each annotation includes the object ID, class label, and bounding box orientation. Keyframe interpolation is used to propagate annotations across temporally adjacent frames where object motion is smooth.
Finally, full 3D bounding boxes are reconstructed by estimating object height and vertical extent from the LiDAR points associated with each BEV annotation across all observations of that object over time.
This BEV-based labeling workflow enables efficient annotation while maintaining global consistency across multiple sensor nodes.

\subsection{Spatial Coverage}

\begin{figure}[htbp]
  \begin{minipage}[t]{0.52\textwidth}
    \centering
    \includegraphics[width=\textwidth]{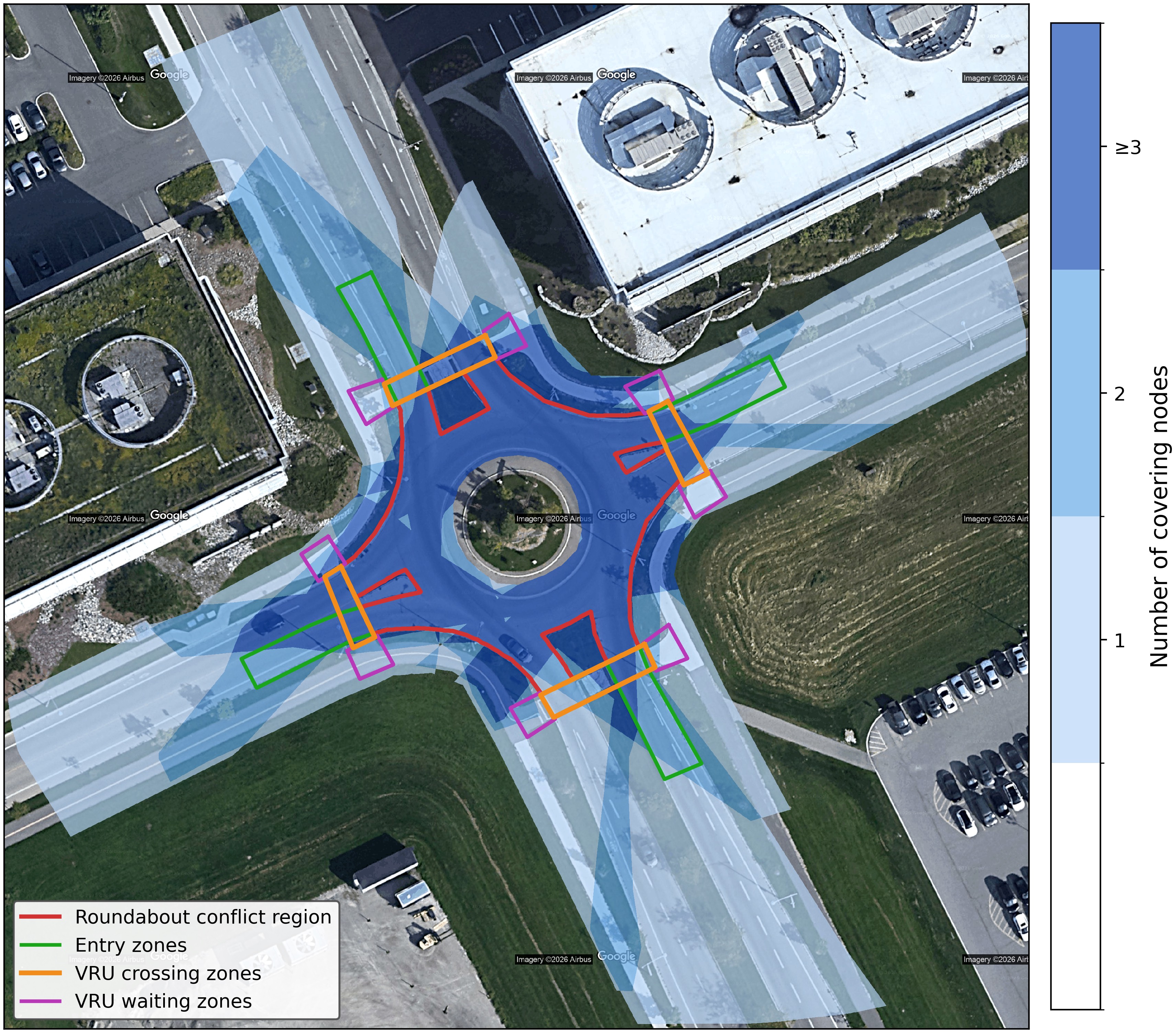}
    \caption{Spatial coverage multiplicity of the eight infrastructure nodes over
    the monitored roundabout site. Color indicates the number of nodes covering
    each location; overlaid boundaries mark the roundabout conflict region, entry
    zones, VRU crossing zones, and VRU waiting zones.}
    \label{Fig:CoverageRegion}
  \end{minipage}
  \hfill
  \begin{minipage}[t]{0.46\textwidth}
    \centering
    \includegraphics[width=\textwidth]{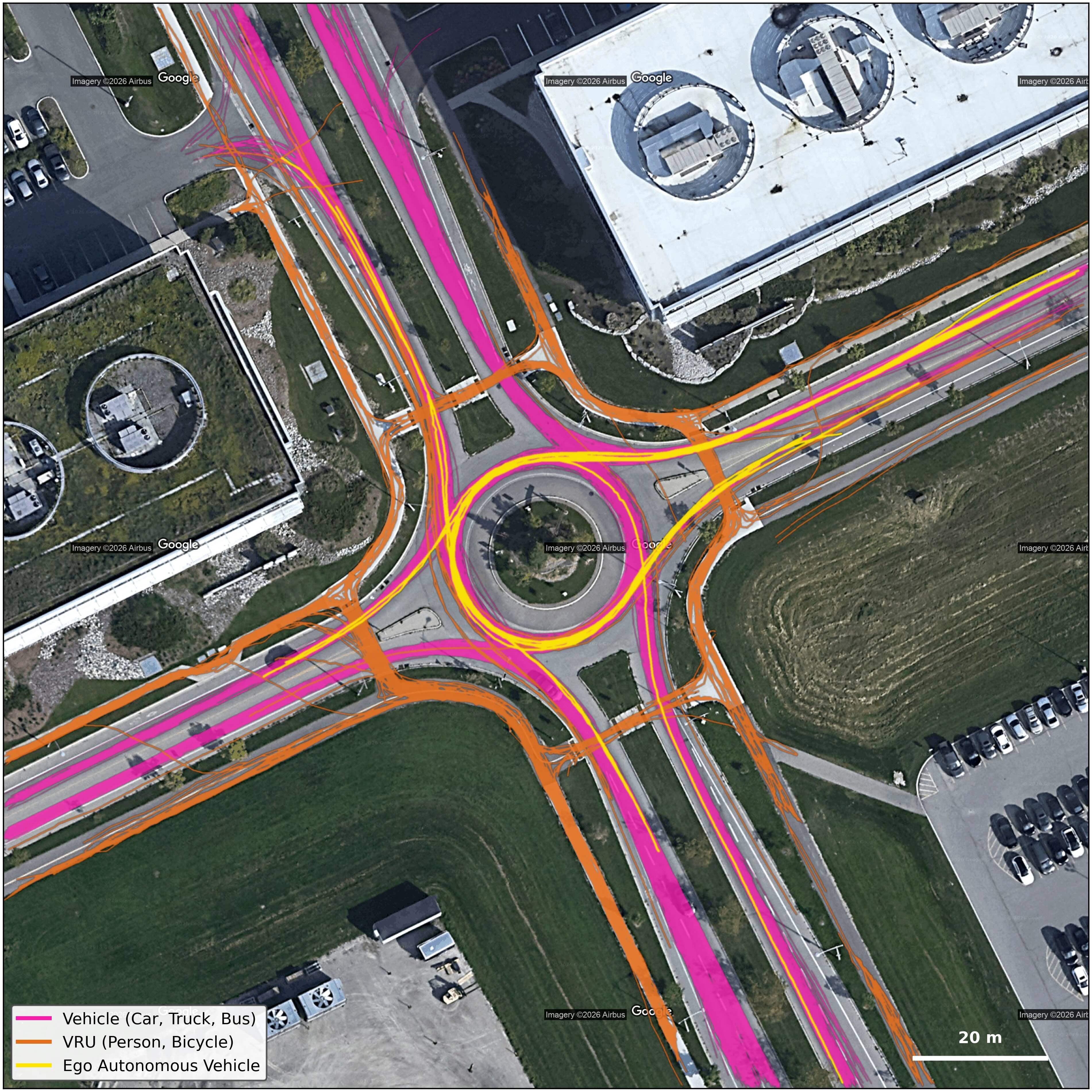}
    \caption{Aggregated trajectories across all scenarios and weather conditions. Magenta denotes other vehicles (Car, Truck, Bus), orange denotes VRUs (Person, Bicycle), and yellow denotes the ego autonomous vehicle (AV).}
    \label{Fig:CoInfraTrajectories}
  \end{minipage}
\end{figure}

The spatial coverage analysis in Fig.~\ref{Fig:CoverageRegion} shows that sensing redundancy is concentrated in the regions most critical for transportation safety. While 96.2\% of the total drivable area is covered by at least one node, all conflict zones are fully covered by at least one node. In particular, the roundabout ring—where merging, yielding, and trajectory conflicts are most frequent—achieves 99.5\% coverage by at least two nodes and 96.5\% by at least three, with a mean coverage multiplicity of 3.83. VRU crossing zones also maintain strong overlap, with 92.4\% dual-node and 67.4\% triple-node coverage. By contrast, waiting zones show lower overlap (45.7\% dual-node), consistent with a deployment that prioritizes active crossing paths over staging areas.



\subsection{Dataset Statistics}



\begin{figure*}[t]
    \centering
    \begin{subfigure}[t]{0.245\textwidth}
        \centering
        \includegraphics[width=\linewidth]{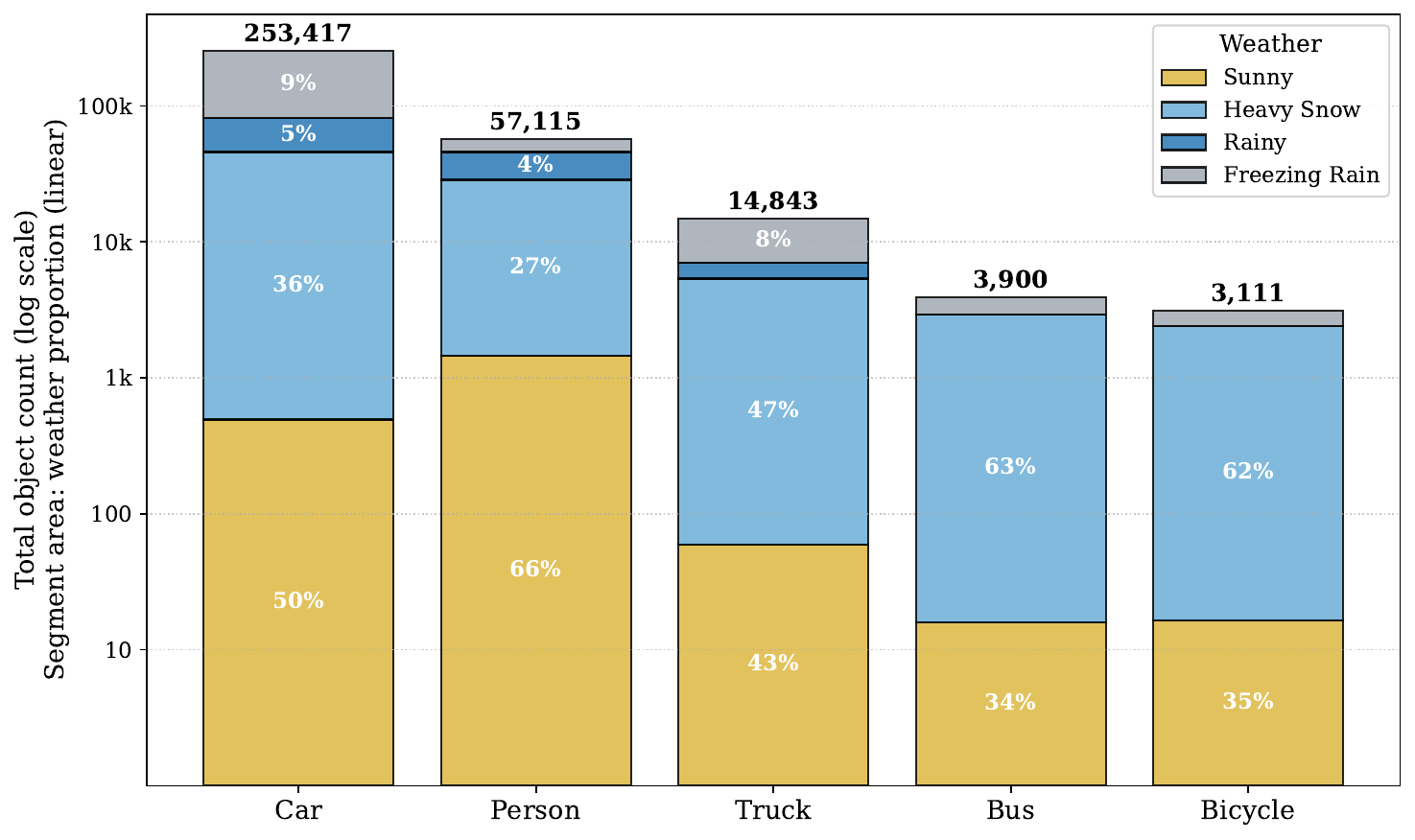}
        \caption{Distribution of annotated objects in the global frame.}
        \label{Fig:ObjectCounts}
    \end{subfigure}\hfill
    \begin{subfigure}[t]{0.245\textwidth}
        \centering
        \includegraphics[width=0.9\linewidth]{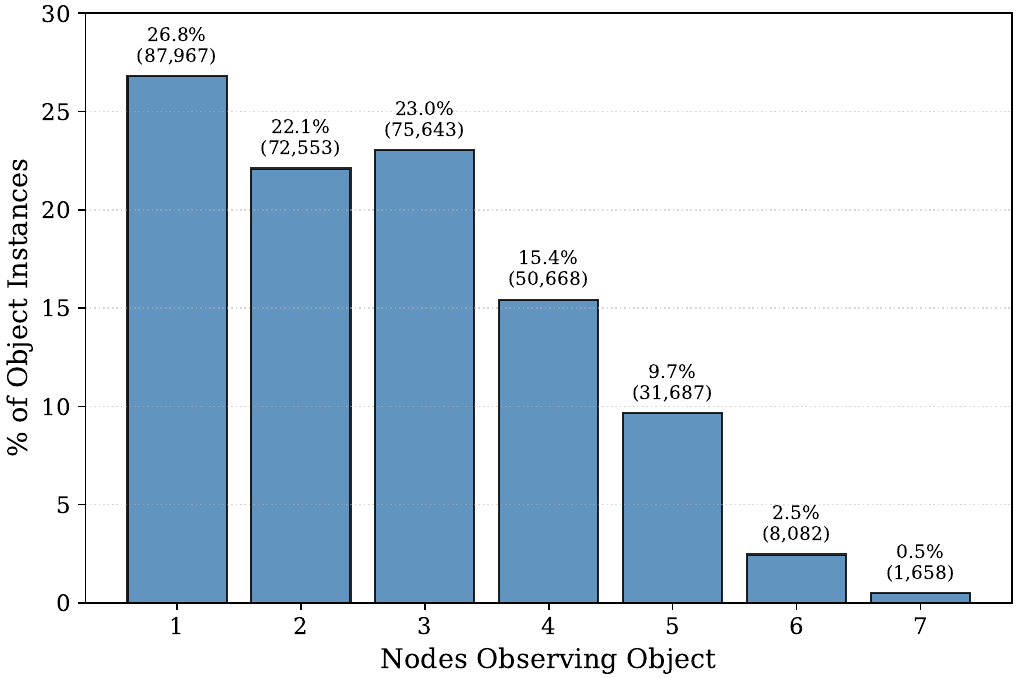}
        \caption{Distribution of the number of infrastructure nodes observing the same object.}
        \label{Fig:ObjectOverlap}
    \end{subfigure}\hfill
    \begin{subfigure}[t]{0.245\textwidth}
        \centering
        \includegraphics[width=\linewidth]{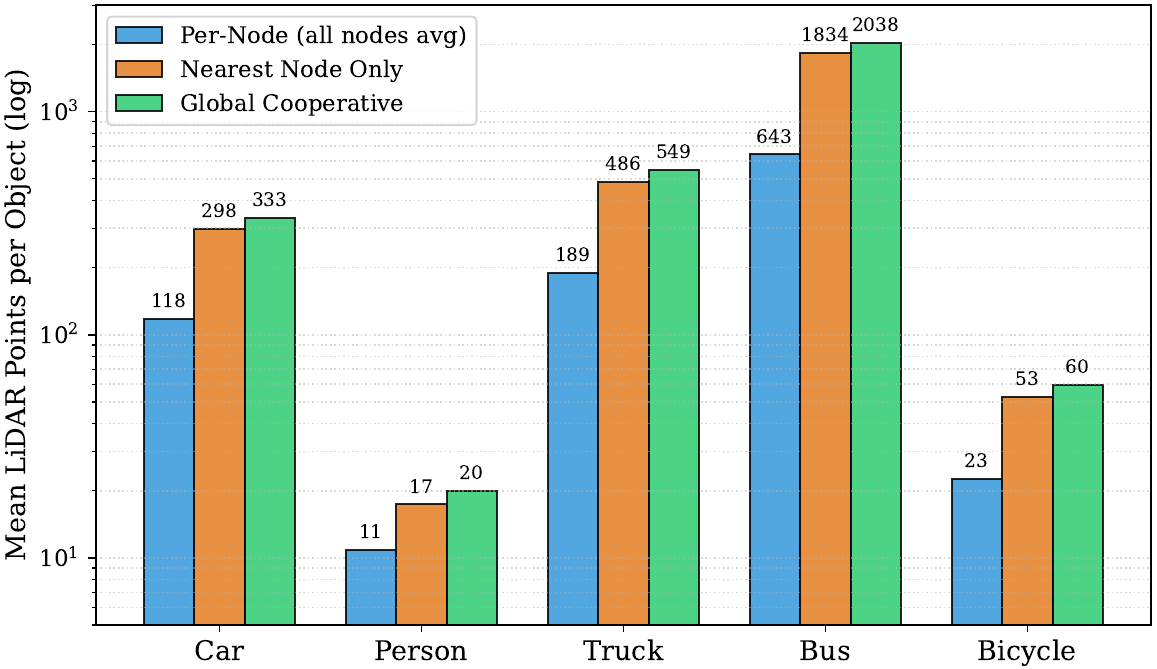}
        \caption{Comparison of LiDAR point density for single-node and multi-node fusion.}
        \label{Fig:single_vs_coop_lidar_density}
    \end{subfigure}\hfill
    \begin{subfigure}[t]{0.245\textwidth}
        \centering
        \includegraphics[width=\linewidth]{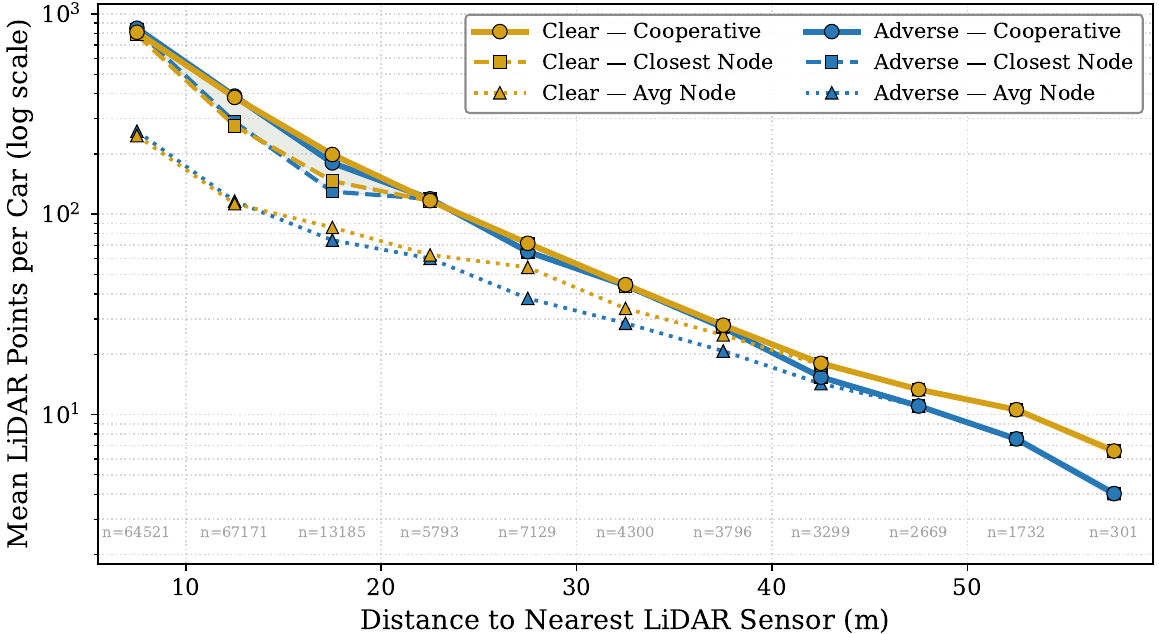}
        \caption{Mean LiDAR points per car under clear and adverse weather conditions.}
        \label{Fig:clear_vs_adverse_lidar_density}
    \end{subfigure}
    \caption{Dataset statistics and LiDAR point-density characteristics of CoInfra.}
    \label{fig:dataset_stats_density}
\end{figure*}

The spatial distribution of all trajectories from CoInfra-I2I and CoInfra-V2I is
visualized in Fig.~\ref{Fig:CoInfraTrajectories}, confirming dense coverage of all
four roundabout approach legs and the central conflict ring.
The primary CoInfra-I2I subset contains 294k LiDAR frames and 589k camera images collected from the eight-node deployment. In total, the dataset includes more than 332k unique 3D bounding boxes in the global coordinate frame, as shown in Fig.~\ref{Fig:ObjectCounts}. 


A defining property of the dataset is the degree of multi-node observation overlap: each object is observed by 2.68 nodes on average, and over 73\% of objects are visible from at least two nodes (Fig.~\ref{Fig:ObjectOverlap}). 
This overlap is by design: nodes are densely deployed around conflict-critical regions, which directly supports the cooperative fusion and observability analyses in Section~\ref{sec:experiments}.
Cooperative sensing also substantially increases per-object LiDAR point density compared with single-node observations, particularly for small and distant objects (Fig.~\ref{Fig:single_vs_coop_lidar_density}).

Object point density also varies significantly across classes. Large vehicles such as buses may contain over one thousand LiDAR points per instance, whereas pedestrians often contain fewer than ten points. This sparsity highlights the importance of cooperative multi-node sensing and map-aware perception for reliable VRU detection.

Finally, adverse weather conditions further reduce LiDAR point density. Fig.~\ref{Fig:clear_vs_adverse_lidar_density} shows that point density decreases significantly under adverse conditions compared with clear weather, particularly at longer sensing distances.

\subsection{Example Data}
\begin{figure*}[h]
  \centering
  \includegraphics[width=0.95 \textwidth]{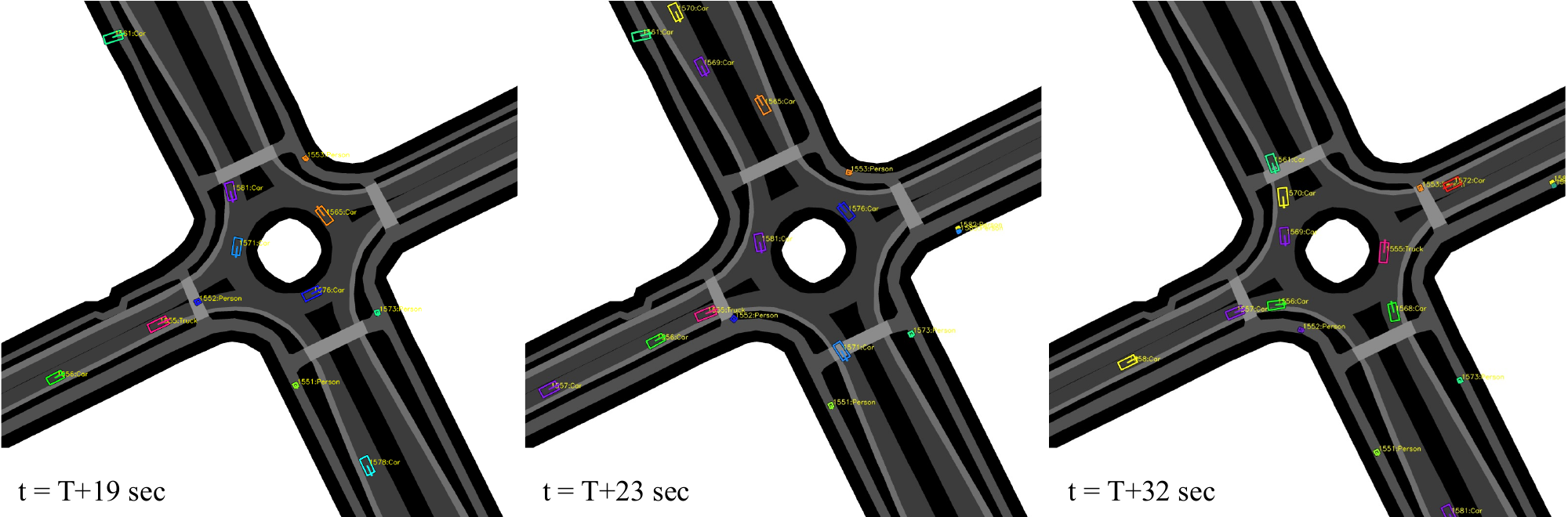}
  \caption{ \textbf{Examples showing object interactions in CoInfra.}
  BEV sequences at three time points illustrate dynamic behaviors in the roundabout, such as vehicles merging, yielding, and interacting with other road users. \textit{T} means the start of this sequence
  November 20, 2025, at 17:03:56. }
  \label{Fig: Interaction}
\end{figure*}

\begin{figure*}[htbp]
  \centering
  \includegraphics[width=0.95 \textwidth]{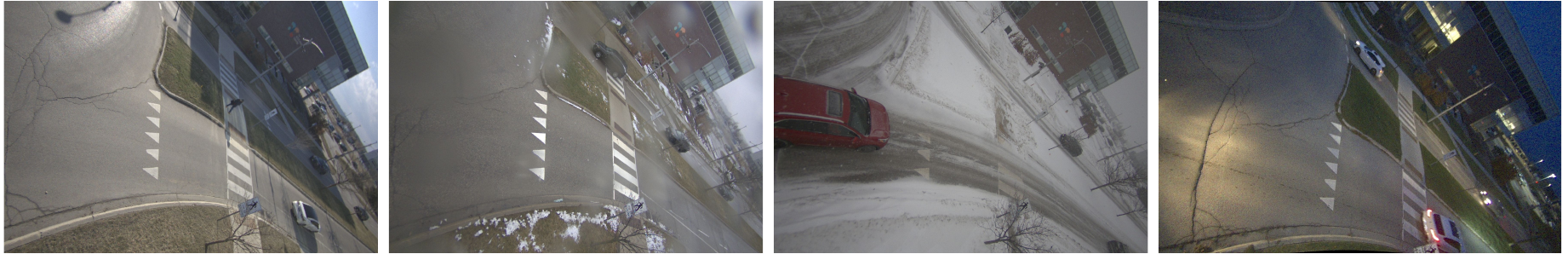}
  \caption{\textbf{Examples showing diverse environmental conditions in
  CoInfra} including Sunny, Rainy, Snowy and Night.}
  \label{Fig: ExampleWeather}
\end{figure*}



\begin{figure*}[t]
  \centering
  \begin{subfigure}[t]{0.48\textwidth}
    \centering
    \includegraphics[width=\linewidth]{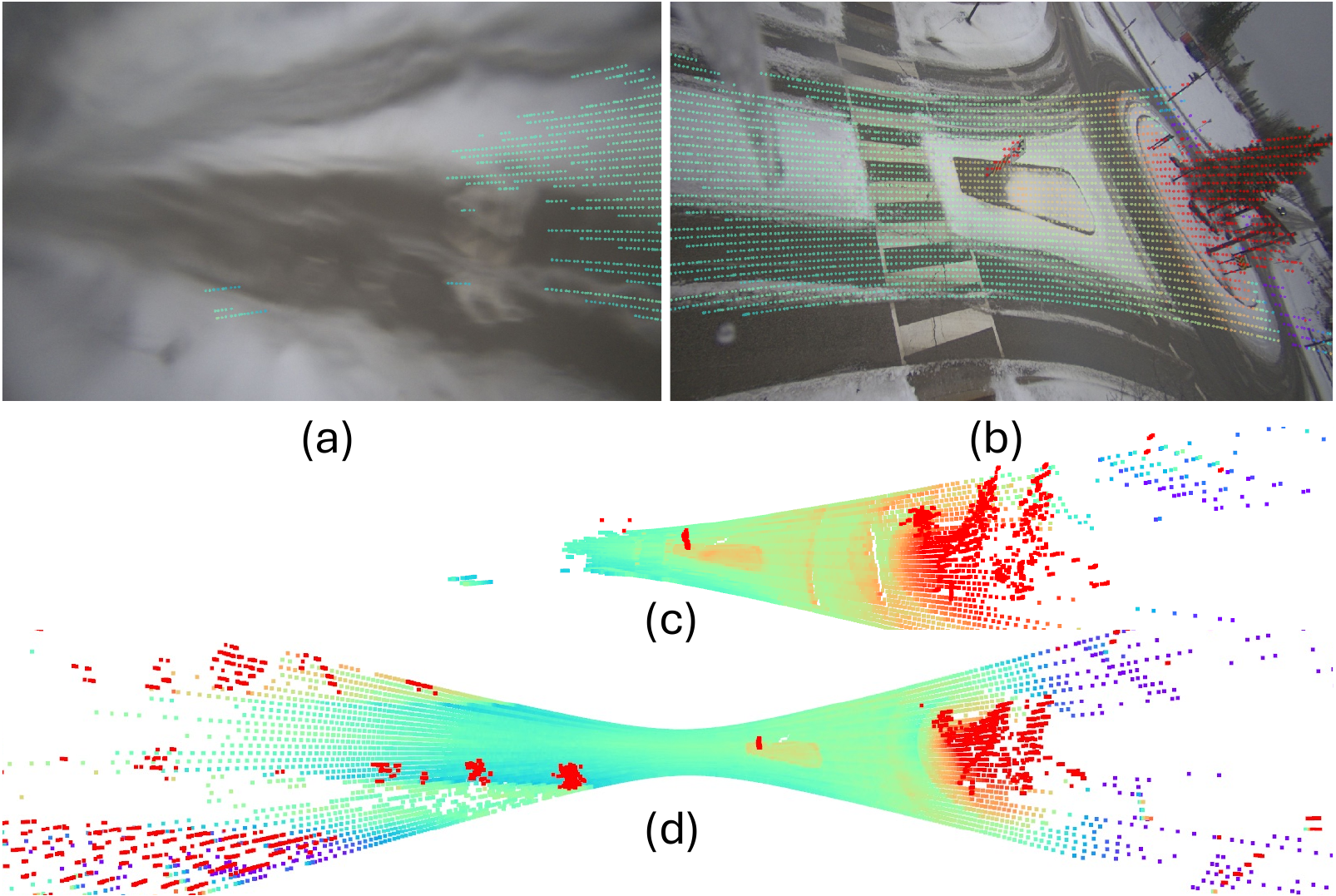}
    \caption{\textbf{Freezing rain example from CoInfra.} (a) and (b): Left and right camera images under freezing rain, with significant blur from ice accumulation. (c): LiDAR point cloud from the same period, showing regions with missing data (NaNs) due to ice blocking the beams. (d): LiDAR point cloud under sunny conditions, showing normal returns for comparison.}
    \label{Fig:ExampleFreezingRain}
  \end{subfigure}\hfill
  \begin{subfigure}[t]{0.49\textwidth}
    \centering
    \includegraphics[width=\linewidth]{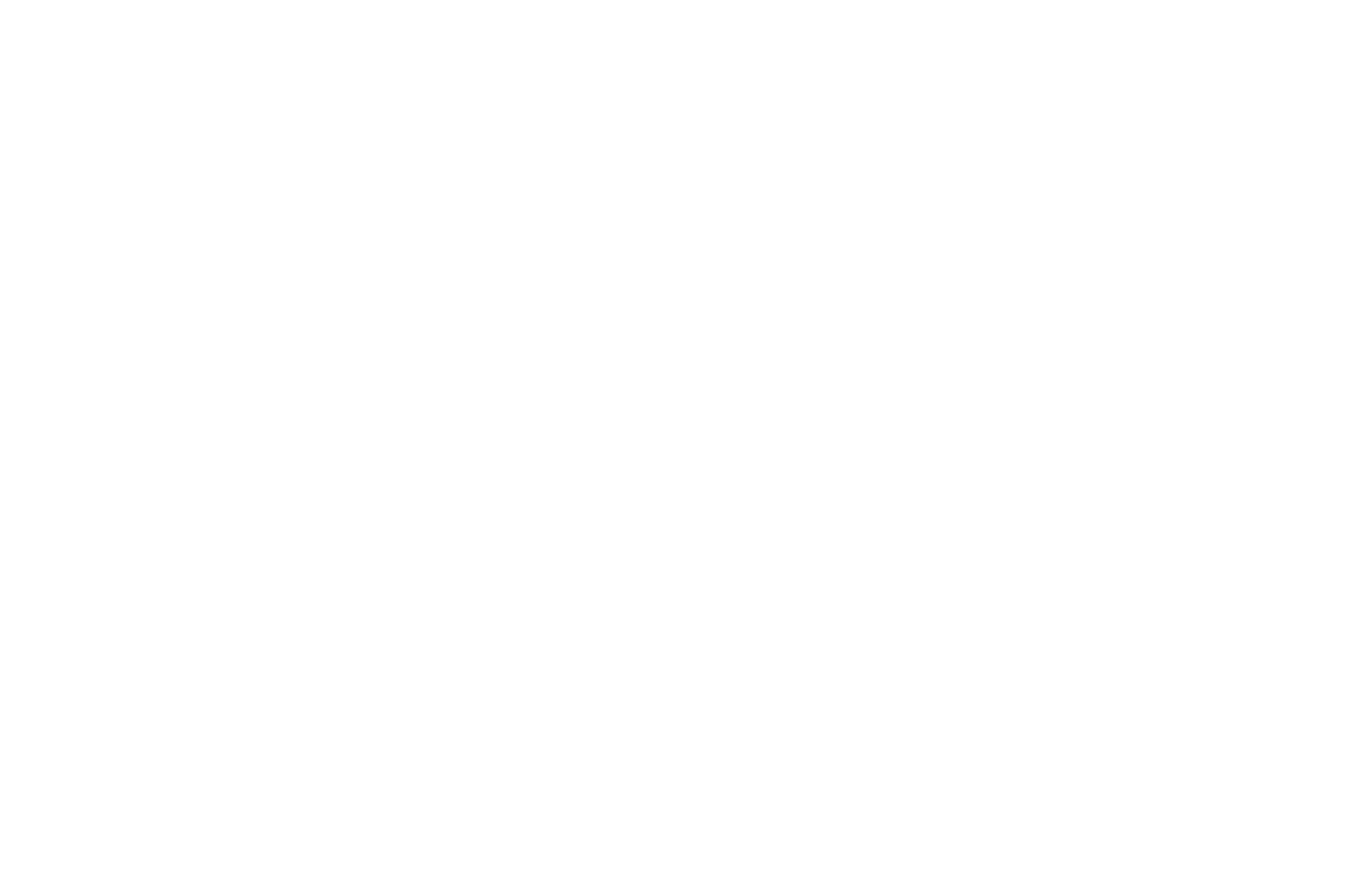}
    \caption{\textbf{Heavy snow example from CoInfra.} Left and right camera images and the corresponding LiDAR point cloud under heavy snow. The LiDAR view is shown in the sensor frame, and increased noise points can be observed, especially near the sensor.}
    \label{Fig:ExampleSnow}
  \end{subfigure}
  \caption{\textbf{Adverse weather examples from CoInfra.} Representative freezing-rain and heavy-snow cases illustrating severe visual degradation and LiDAR noise under challenging real-world conditions.}
  \label{Fig:ExampleAdverseWeather}
\end{figure*}

To illustrate the breadth of scenarios captured by CoInfra, we present representative examples spanning multi-node multi-modal perception, traffic interactions, and adverse weather in Fig.~\ref{fig:dataset_example}, Fig.~\ref{Fig: Interaction}, Fig.~\ref{Fig: ExampleWeather} and Fig.~\ref{Fig:ExampleAdverseWeather}.
Fig.~\ref{fig:dataset_example} shows a heavy-snow scene from the multi-node deployment and illustrates a key property of the dataset: globally consistent scene understanding despite weather degradation, occlusion, and fragmented viewpoints. The complementary views across nodes help preserve object coverage even when individual sensors are degraded.
Fig.~\ref{Fig: Interaction} highlights the interaction diversity captured in the monitored roundabout, including vehicle merging, yielding, and interactions with vulnerable road users. These conflict-rich scenarios make the dataset useful not only for cooperative detection, but also for tracking and behavior-aware evaluation.
Fig.~\ref{Fig: ExampleWeather} shows representative samples under sunny, rainy, snowy, and night conditions, demonstrating the environmental diversity of the dataset.
Fig.~\ref{Fig:ExampleFreezingRain} presents a particularly challenging freezing-rain case, where ice accumulation severely degrades both camera image quality and LiDAR returns. Fig.~\ref{Fig:ExampleSnow} shows a heavy-snow scene, where reduced visibility and increased near-sensor LiDAR noise further complicate perception. 
Together, these examples show that CoInfra captures the environmental variability, sensing degradation, and interaction complexity needed to study cooperative perception under realistic deployment conditions.

\section{Experiments}
\label{sec:experiments}

This section evaluates CoInfra from systems feasibility to transportation value. We first characterize the real-world 5G communication envelope of the deployed platform (Section~\ref{sec:5g_analysis}). We then evaluate delay-aware synchronization under recorded network latency (Section~\ref{sec:sync_experiments}). Next, we establish cooperative perception benchmarks on CoInfra-I2I to quantify the detection gains provided by multi-node sensing (Section~\ref{sec:perception_benchmarks}). Finally, we show that these gains translate into the paper's main transportation-facing outcome: improved ego-centric observability of safety-critical participants around a connected autonomous vehicle (Section~\ref{sec:v2i_case_study}).

\subsection{Real-World 5G Communication Analysis}
\label{sec:5g_analysis}

We evaluate the impact of real-world communication on cooperative perception using 5G latency and reliability measurements collected from our 14-node deployment under sunny, rainy, and snowy conditions.

End-to-end latency is decomposed into local processing latency $\delta^{\text{proc}}$ (sensor acquisition to perception output on the Jetson Orin NX) and communication latency $\delta^{\text{comm}}$ (message transmission from node to cloud). Each message is timestamped at both the edge and the cloud, enabling precise per-message latency measurement. The local processing latency $\delta^{\text{proc}}$ is relatively stable, with mean 52.1~ms and standard deviation 4.1~ms~\cite{yang2025real}. The remainder of this subsection focuses on the 5G communication latency $\delta^{\text{comm}}$.


\subsubsection{Latency Distribution Under Different Payload}

\begin{table}[t]
\centering
\caption{Communication latency under different payload types. Payload sizes are reported per node per timestamp.}
\label{tab:payload_latency}
\scriptsize
\begin{tabular}{lcccccc}
\toprule
Payload & Size & Mean (ms) & Median (ms) & P95 (ms) & Drop rate (\%)\\
\midrule
Bounding boxes & 1--2 KB & 24.2 & 19.8 & 38.2 & 0.00\\
Cropped point cloud & 30--40 KB & 44.3 & 44.6 & 64.7 & 0.00\\
Two compressed images & 500--600 KB & 2189.6 & 2151.8 & 4266.4 & 63.75\\
\bottomrule
\end{tabular}
\end{table}

To quantify how payload size affects communication performance, we measured 5G transmission latency under three representative message types: object-level detections (bounding boxes), cropped point clouds after ground removal, and compressed camera images. Table~\ref{tab:payload_latency} summarizes the results.

Lightweight object-level communication achieves the best performance, with a mean latency of 24.2~ms and P95 of 38.2~ms for 1--2~KB bounding box messages. Increasing the payload to cropped point clouds (30--40~KB) results in moderate latency growth (mean 44.3~ms, P95 64.7~ms) while maintaining zero packet loss. In contrast, transmitting two compressed images (500--600~KB) leads to severe network congestion, with mean latency exceeding 2.19~s and a packet drop rate of 63.75\%. 

These results define a clear communication hierarchy for deployable cooperative perception. Compact object-level outputs are well suited to real-time multi-node operation. Moderately richer geometric payloads, such as cropped point clouds, remain feasible but incur noticeably higher delay. Raw image transmission, however, does not scale in the multi-node setting. Overall, the measurements support the design choice of transmitting structured perception outputs rather than raw sensor streams.

\subsubsection{Latency Distributions Under Varying Weather and Network Load}
\begin{table}[t]
\centering
\caption{5G communication latency statistics under different weather and network load conditions.}
\label{tab:5g_weather_latency}
\scriptsize
\begin{tabular}{lcccccccc}
\toprule
Condition & Mean & $\sigma$ & Median & P95 & P99 & $>$60 ms & $>$100 ms & $>$200 ms\\
 & (ms) & (ms) & (ms) & (ms) & (ms) & (\%) & (\%) & (\%) \\
\midrule
Clear (Busy)  & 32.2 & 9.4  & 32.0 & 43.1 & 49.8 & 0.443 & 0.162 & 0.042 \\
Clear (Idle)  & 28.8 & 8.0  & 29.1 & 40.4 & 42.6 & 0.182 & 0.083 & 0.007 \\
Rainy (Busy)  & 32.4 & 11.8 & 31.4 & 44.2 & 51.4 & 0.590 & 0.308 & 0.109 \\
Rainy (Idle)  & 28.2 & 9.7  & 28.0 & 38.9 & 43.0 & 0.201 & 0.133 & 0.049 \\
Snowy (Busy)  & 31.7 & 13.0 & 30.3 & 44.3 & 54.1 & 0.801 & 0.421 & 0.139 \\
Snowy (Idle)  & 28.7 & 7.6 & 28.2 & 41.2 & 47.0 & 0.223 & 0.064 & 0.007 \\
\bottomrule
\end{tabular}
\end{table}

To evaluate the impact of environmental conditions on communication performance, we conducted latency measurements under clear, rainy, and snowy weather, with two network load regimes: \textit{busy} (10:00--16:00) and \textit{idle} (01:00--05:00). During the experiment, each node transmitted Autoware perception messages at 10~Hz containing bounding boxes, tracking attributes, and downsampled point cloud clusters. Payload sizes were randomized between 1--12~KB to emulate realistic perception outputs. Table~\ref{tab:5g_weather_latency} summarizes the measured latency statistics.

Across all conditions, the mean latency remains stable at approximately 28--32~ms. Weather has little influence on average latency: clear, rainy, and snowy conditions show nearly identical mean delays. However, adverse weather slightly increases tail latency; for example, the P99 latency increases from 49.8~ms under clear conditions to 54.1~ms during snow, and the fraction of messages exceeding 60~ms increases from 0.443\% to 0.801\%.

Network load has a larger impact than weather. Idle periods consistently exhibit lower mean latency ($\sim$28~ms) and tighter distributions, while busy periods show slightly higher averages ($\sim$32~ms) and heavier tails.
Even under busy conditions and adverse weather, the P95 latency for compact payloads remains below 45~ms and the fraction of messages exceeding 100~ms stays below 0.5\%. Having established the communication envelope, the next section addresses how to exploit it for timely multi-node fusion.

\subsection{Delay-Aware Synchronization Under Recorded 5G Latency}
\label{sec:sync_experiments}

We evaluate the delay-aware synchronization protocol through two experiments: simultaneous sensor triggering accuracy and adaptive fusion-window performance under recorded 5G delays.

\subsubsection{Simultaneous Sensor Triggering}

\begin{figure*}[t]
\centering

\begin{minipage}[t]{0.53\textwidth}
    \centering
    \begin{subfigure}[t]{0.5\linewidth}
        \centering
        \includegraphics[width=\linewidth]{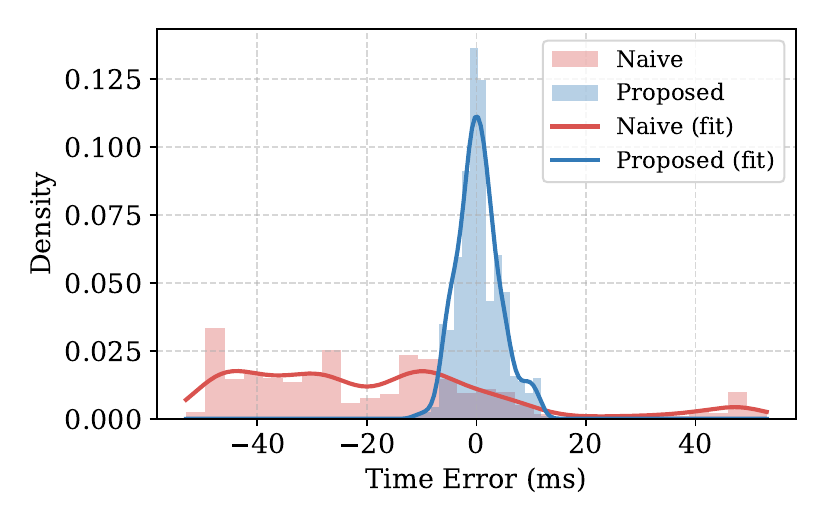}
        \caption{Acquisition-time deviation.}
    \end{subfigure}\hfill
    \begin{subfigure}[t]{0.5\linewidth}
        \centering
        \includegraphics[width=\linewidth]{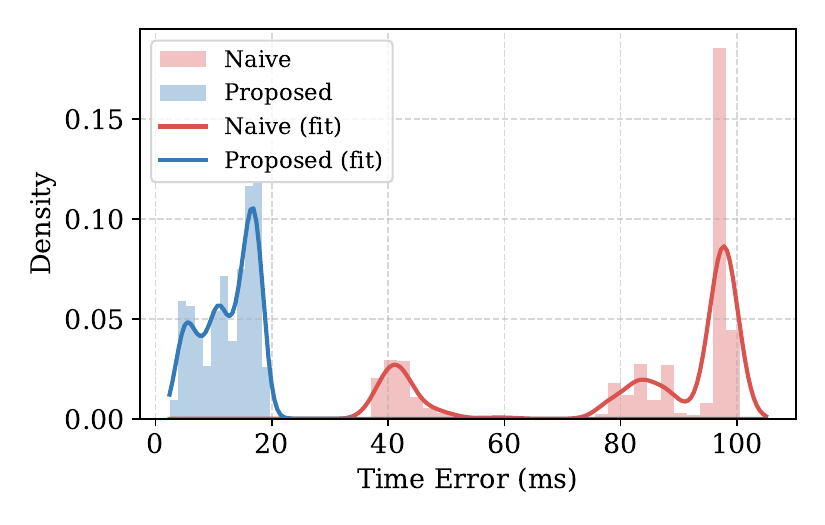}
        \caption{Min--max inter-node acquisition delay.}
    \end{subfigure}
    \captionof{figure}{Evaluation of simultaneous sensor triggering compared with asynchronous acquisition.}
    \label{fig:sync_trigger_eval}
\end{minipage}\hfill
\begin{minipage}[t]{0.46\textwidth}
    \centering
    \includegraphics[width=\linewidth]{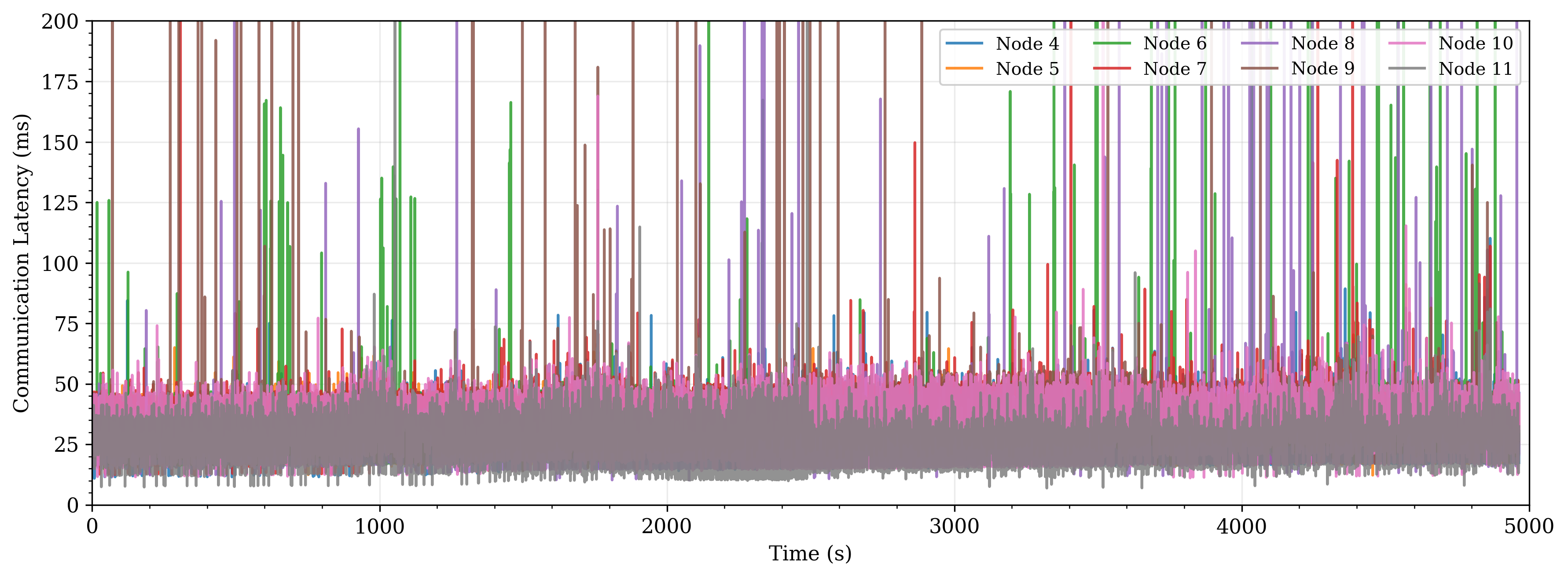}
    \captionof{figure}{Recorded per-node 5G communication latency used in the trace-driven adaptive fusion experiment. Most transmissions arrive within 20--50\,ms, but occasional spikes exceed 100\,ms.}
    \label{fig:real_latency_for_adaptive_fusion}
\end{minipage}

\end{figure*}

We evaluate the effectiveness of the proposed globally aligned sensor triggering mechanism. 
In the deployed system, all infrastructure nodes are triggered at fixed global time anchors spaced at 100~ms intervals, ensuring that LiDAR and camera data are acquired concurrently across the network. 
This is compared with a naive asynchronous configuration in which sensors operate independently without coordinated triggering.
Fig.~\ref{fig:sync_trigger_eval}(a) shows the distribution of acquisition-time deviations from the global time anchor. 
Under simultaneous triggering, the timing errors are tightly concentrated around zero and remain mostly within $\pm$5~ms. 
In contrast, the asynchronous configuration produces substantially larger deviations, reaching approximately $\pm$40~ms.
Fig.~\ref{fig:sync_trigger_eval}(b) presents the min–max inter-node delay for each trigger event, defined as the time difference between the earliest and latest node acquisitions. 
The proposed triggering mechanism limits this delay to below 20~ms, whereas the asynchronous configuration produces delays approaching 100~ms.

\subsubsection{Trace-Driven Evaluation of Adaptive Synchronization}

We evaluate the proposed adaptive synchronization strategy on CoInfra-I2I using \emph{recorded} 5G communication delays from the deployed intersection. For each anchor time $t_k$, the node-level ROI annotations from Nodes~4--11 are treated as local perception outputs, while the corresponding global ROI annotation serves as the reference scene state. This trace-driven design preserves the real multi-node visibility patterns and object motion in CoInfra while enabling controlled comparison of synchronization policies under authentic communication conditions.

\textbf{Setup.}
Each local message is assigned a recorded per-node delay sample from the deployed 5G network, so every evaluation frame uses an observed rather than synthetic communication profile. As shown in Fig.~\ref{fig:real_latency_for_adaptive_fusion}, most transmissions arrive within 20--50\,ms, but occasional per-node spikes above 100\,ms are also present, and these tail events are the main challenge for cloud-side synchronization. At each anchor time, each node contributes the objects visible in its ROI, and local perception uncertainty is emulated by adding zero-mean Gaussian position noise with $\sigma_p=0.20$\,m. Each node also maintains a constant-velocity EKF that can optionally predict temporarily delayed tracks. The cloud computes the adaptive fusion deadline online as Eq.~\ref{eq:fusion_window}.
We compare four synchronization strategies. \textit{Wait-for-all} releases fusion only after all node messages for the current anchor have arrived. \textit{Adaptive $2\sigma$ without EKF} uses the online deadline $\max_i(\hat{\mu}_i + 2\hat{\sigma}_i)$ and fuses only the messages that arrive before that time. \textit{Adaptive $2\sigma$ with EKF} uses the same deadline, but substitutes missing expected node contributions with local EKF predictions from the most recent observation. \textit{Adaptive $3\sigma$ with EKF} uses the same mechanism with the more conservative deadline $\max_i(\hat{\mu}_i + 3\hat{\sigma}_i)$. In all cases, contributions are associated by track ID and fused by inverse-variance weighted averaging, with predicted states assigned inflated covariance. The fused result is then propagated to the actual release time and evaluated against interpolated ground truth at that time.

\textbf{Metrics.}
We report reaction time from anchor to fusion release, full-arrival rate, object-level completeness, track recall, and output-time position error for all objects, vehicles, and VRUs.

\begin{table*}[t]
\centering
\caption{Trace-driven evaluation of synchronization strategies on CoInfra using recorded 5G communication latency. The adaptive deadline is computed online as $\max_i(\hat{\mu}_i+n\hat{\sigma}_i)$. \textbf{Bold} indicates the best value in each column.}
\label{tab:sync_eval_real}
\scriptsize
\setlength{\tabcolsep}{4pt}
\begin{tabular}{l c ccc ccc ccc}
\toprule
 & & \multicolumn{3}{c}{Reaction time (ms)} &
   \multicolumn{3}{c}{Completeness} &
   \multicolumn{3}{c}{Output-time position error (m)} \\
\cmidrule(lr){3-5} \cmidrule(lr){6-8} \cmidrule(lr){9-11}
Method &
Deadline &
Mean & P95 & P99 &
Full arr.\ (\%) & Obj.\ (\%) & Recall (\%) &
All & Veh. & VRU \\
\midrule
Wait-for-all
  & ---
  & 44.1 & 49.8 & 75.7
  & \textbf{100.0} & \textbf{100.0} & \textbf{99.41}
  & 0.193 & 0.191 & 0.202 \\[2pt]
Adaptive ($2\sigma$, no EKF)
  & 55.9
  & \textbf{42.4} & \textbf{49.7} & \textbf{57.0}
  & 97.9 & 99.7 & 99.34
  & \textbf{0.187} & \textbf{0.184} & \textbf{0.200} \\[2pt]
Adaptive ($2\sigma$, +EKF)
  & 55.9
  & \textbf{42.4} & \textbf{49.7} & \textbf{57.0}
  & 97.9 & \textbf{100.0} & \textbf{99.41}
  & \textbf{0.187} & \textbf{0.184} & \textbf{0.200} \\[2pt]
Adaptive ($3\sigma$, +EKF)
  & 65.8
  & 42.5 & 49.8 & 63.2
  & 98.9 & \textbf{100.0} & \textbf{99.41}
  & \textbf{0.187} & \textbf{0.184} & \textbf{0.200} \\
\bottomrule
\end{tabular}
\end{table*}

\textbf{Results and discussion.}
Table~\ref{tab:sync_eval_real} shows that adaptive synchronization primarily improves tail latency rather than mean latency. Wait-for-all achieves perfect completeness by construction, but its P99 reaction time reaches 75.7\,ms because fusion is repeatedly gated by the slowest node. Adaptive $2\sigma$ reduces P99 to 57.0\,ms while also slightly improving mean reaction time. Without state prediction, the earlier deadline causes a small loss in completeness; with EKF compensation, however, adaptive $2\sigma$ recovers full object completeness and the same recall as wait-for-all, while retaining the lower output delay and slightly reducing position error (0.187\,m vs.\ 0.193\,m). Increasing the deadline to $3\sigma$ improves full-arrival rate but yields no meaningful gain in recall or localization accuracy. Under the measured latency regime, adaptive $2\sigma$ with EKF therefore provides the best operating point.
With the communication and synchronization envelopes now characterized, the next question is whether multi-node cooperation actually improves scene understanding.

\subsection{Cooperative 3D Object Detection Benchmarks}
\label{sec:perception_benchmarks}

We establish baseline benchmarks for cooperative 3D object detection on CoInfra-I2I to demonstrate the dataset utility and provide reproducible reference results.

\subsubsection{Experimental Setup}
The dataset is randomly split into training (80\%), validation (10\%), and test (10\%) sets at the global anchor timestamp level. 
We benchmark six baselines formed by crossing three sensing modalities (camera-only, LiDAR-only, LiDAR--camera) with two cooperative settings: \emph{global fusion}, in which all node observations are projected into a shared world-aligned BEV before detection, and \emph{local fusion}, in which each node performs detection independently and the resulting detections are merged in the common ground frame via cross-node post-processing. All BEV representations are registered to the global HD map, providing spatial context for drivable-area reasoning. This yields six baselines: \textit{GlobalCam}, \textit{GlobalLidar}, \textit{GlobalLidarCam}, \textit{LocalCam}, \textit{LocalLidar}, and \textit{LocalLidarCam}. We report AP@0.5 for three categories: \textbf{Bus}, \textbf{LV} (light vehicles, which include car and truck), and \textbf{VRU} (person and bicycle), further stratified by distance to the closest infrastructure node.

\subsubsection{Results and Analysis}

\begin{figure*}[t]
  \centering
  \includegraphics[width=0.95\linewidth]{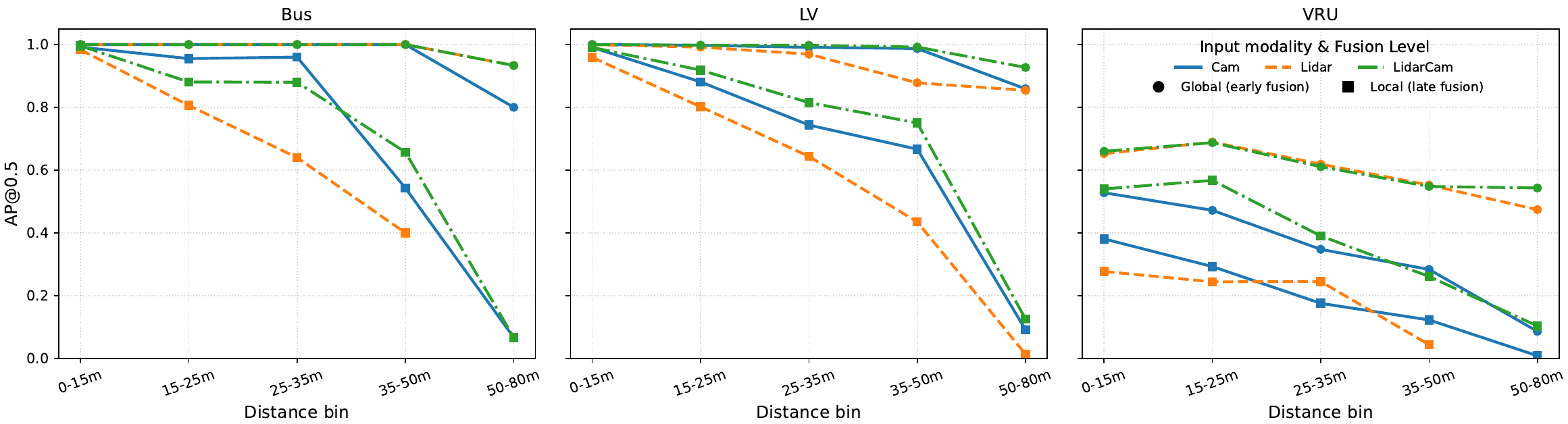}
  \caption{AP@0.5 vs.\ distance to the closest node (Bus/LV/VRU subplots). Markers encode fusion level (Global vs.\ Local); line color/style encodes input modality (Cam vs.\ LiDAR vs.\ LiDAR+Cam).}
  \label{fig:distance_ap50}
\end{figure*}

\textbf{Distance-dependent performance.}
Fig.~\ref{fig:distance_ap50} shows AP@0.5 as a function of distance to the closest node. Global fusion consistently outperforms local fusion across all classes, with the largest margin observed for VRUs at longer range. This trend is consistent with the geometric property of the dataset: cooperative sensing substantially increases the number of LiDAR returns per object, especially for sparse classes. Averaged over all nodes, the mean number of LiDAR points per object is 11 for \textit{Person} and 23 for \textit{Bicycle}, while global cooperative aggregation increases them to 20 and 60, respectively. Even for larger objects, cooperative fusion still improves point density, e.g., from 118 to 333 for \textit{Car} and from 643 to 2038 for \textit{Bus}.

\textbf{Weather robustness and safety-relevant regions.}
Table~\ref{tab:weather_region_ap50} reports AP@0.5 across weather conditions and safety-relevant subregions. Three findings stand out. First, \emph{global cooperative models dominate local ones for VRU detection}, and the gap is largest precisely where it matters most: in VRU crossing zones, GlobalLidar achieves 0.720 and GlobalLidarCam 0.726, compared with 0.320 and 0.572 for their local counterparts. Second, \emph{LiDAR-based global models are especially strong in conflict-critical regions}: in the roundabout ring, GlobalLidar and GlobalLidarCam reach near-perfect vehicle AP (0.997--1.000), while local models plateau at 0.930--0.984. Third, \emph{adverse weather degrades VRU detection more than vehicle detection}, with VRU AP dropping from 0.654 (sunny) to 0.585 (snow) for GlobalLidarCam, motivating the multi-node point-density advantage that cooperative sensing provides. Scene-level averages alone understate these effects: the most transportation-relevant gains appear in the interaction-critical subregions where single-node sensing is weakest.

\begin{table*}[t]
\centering
\caption{Overall and region-conditioned AP@0.5. The first three columns report AP@0.5 under all-weather, sunny, and heavy-snow conditions. The remaining columns report AP@0.5 in safety-relevant subregions. Vehicle-related regions (Entry and Ring) are reported for LV and Bus, while pedestrian-related regions (VRUCross and VRUWait) are reported for VRU. ``--'' indicates that the region-class pair is not evaluated.}
\label{tab:weather_region_ap50}
\scriptsize
\setlength{\tabcolsep}{5pt}
\begin{tabular}{llccccccc}
\toprule
\multicolumn{2}{c}{} & \multicolumn{3}{c}{Weather AP@0.5} & \multicolumn{4}{c}{Region-conditioned AP@0.5} \\
\cmidrule(lr){3-5} \cmidrule(lr){6-9}
Method & Class & All & Sunny & Snow & Entry & Ring & VRUCross & VRUWait \\
\midrule
\multirow{3}{*}{GlobalCam}
 & Bus & 0.990 & 0.988 & 0.991 & 0.994 & 0.981 & --    & --    \\
 & LV  & 0.992 & 0.994 & 0.990 & 0.995 & 0.997 & --    & --    \\
 & VRU & 0.404 & 0.447 & 0.356 & --    & --    & 0.483 & 0.428 \\
\midrule
\multirow{3}{*}{LocalCam}
 & Bus & 0.859 & 0.840 & 0.865 & 0.983 & 0.934 & --    & --    \\
 & LV  & 0.882 & 0.896 & 0.875 & 0.972 & 0.983 & --    & --    \\
 & VRU & 0.182 & 0.186 & 0.200 & --    & --    & 0.335 & 0.202 \\
\midrule
\multirow{3}{*}{GlobalLidar}
 & Bus & 0.997 & 0.998 & 0.996 & 1.000 & 1.000 & --    & --    \\
 & LV  & 0.977 & 0.986 & 0.974 & 0.995 & 0.997 & --    & --    \\
 & VRU & 0.615 & 0.657 & 0.571 & --    & --    & 0.720 & 0.635 \\
\midrule
\multirow{3}{*}{LocalLidar}
 & Bus & 0.803 & 0.770 & 0.814 & 0.980 & 0.935 & --    & --    \\
 & LV  & 0.807 & 0.815 & 0.805 & 0.960 & 0.930 & --    & --    \\
 & VRU & 0.128 & 0.120 & 0.158 & --    & --    & 0.320 & 0.193 \\
\midrule
\multirow{3}{*}{GlobalLidarCam}
 & Bus & 0.996 & 0.996 & 0.996 & 1.000 & 0.999 & --    & --    \\
 & LV  & 0.995 & 0.996 & 0.996 & 0.996 & 0.997 & --    & --    \\
 & VRU & 0.618 & 0.654 & 0.585 & --    & --    & 0.726 & 0.651 \\
\midrule
\multirow{3}{*}{LocalLidarCam}
 & Bus & 0.840 & 0.838 & 0.841 & 0.980 & 0.930 & --    & --    \\
 & LV  & 0.905 & 0.916 & 0.904 & 0.986 & 0.984 & --    & --    \\
 & VRU & 0.344 & 0.373 & 0.326 & --    & --    & 0.572 & 0.488 \\
\bottomrule
\end{tabular}
\end{table*}

\textbf{Systems interpretation.}
Taken together, these benchmarks show that cooperative fusion improves detector performance most for sparse and safety-critical objects, but at different communication costs. Combined with the communication analysis in Section~\ref{sec:5g_analysis}, these results reveal a practical trade-off between perception quality and deployment scalability: richer cooperative fusion yields the largest accuracy gains, whereas compact object-level late fusion is more suitable for real-time multi-node deployment. They also suggest that fusion strategy need not be spatially uniform. Late fusion can serve as the default scalable mode, while richer local cues such as cropped point clouds or compact image features may be reserved for high-risk regions such as VRU crossings and waiting zones. Still, detector AP alone does not show whether these gains translate into more complete awareness for a connected vehicle in real traffic interactions. We therefore turn next to ego-centric observability.

\subsection{Transportation-Facing Evaluation on CoInfra-V2I}
\label{sec:v2i_case_study}

The preceding sections established communication feasibility, synchronization timeliness, and cooperative perception gains. We now evaluate whether these capabilities translate into improved situational awareness for a connected autonomous vehicle. In the most demanding ring scenarios, vehicle-only critical-frame completeness is only 33\%--46\%, while combined V2I raises it to 86\%--100\%.

\begin{figure*}[t]
  \centering
  \includegraphics[width=0.99\textwidth]{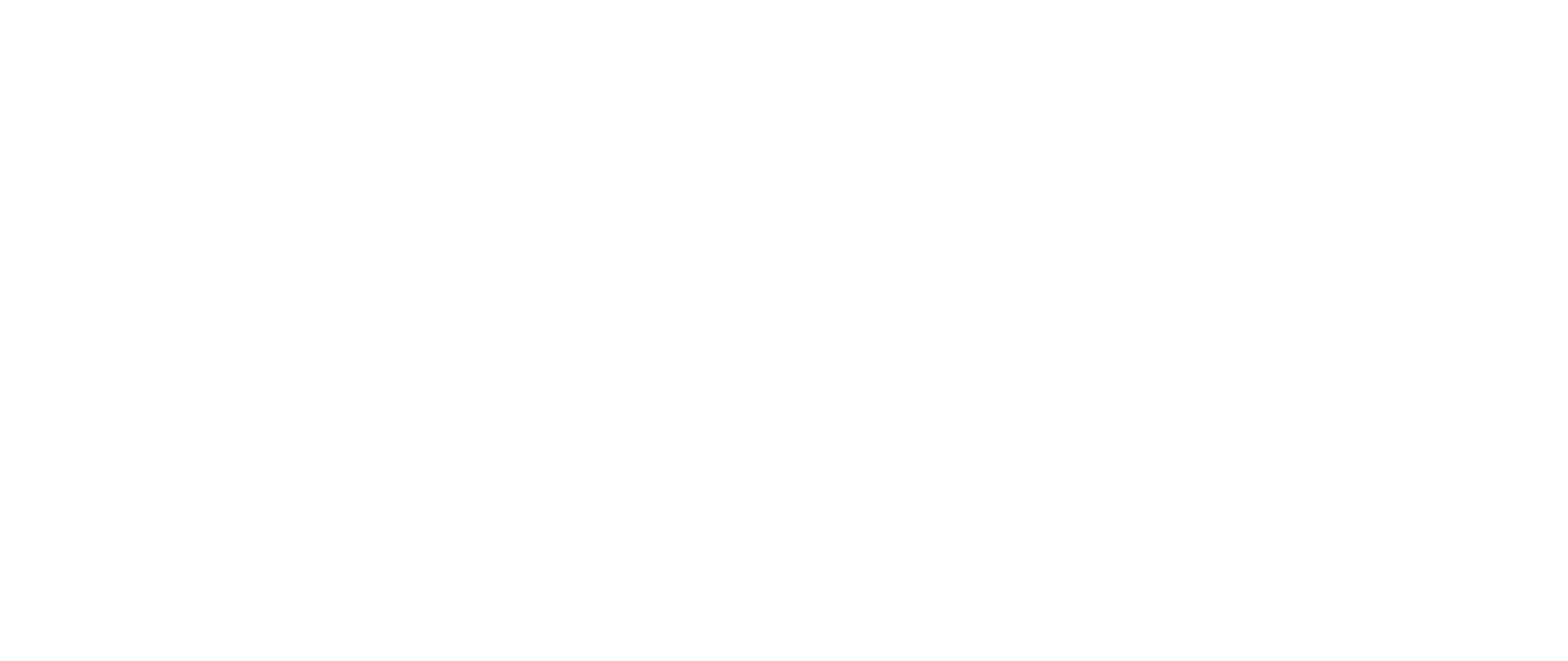}
  \caption{Example of vehicle--infrastructure cooperative perception using the CoInfra-V2I subset. Left: front and rear camera views from the autonomous shuttle. Center: fused perception in Autoware RViz. Right: infrastructure camera and accumulated LiDAR observations. The highlighted region shows a pedestrian occluded from the vehicle view but clearly detected by infrastructure sensors.}
  \label{fig:autoware_integration}
\end{figure*}

Fig.~\ref{fig:autoware_integration} shows a representative case during the autonomous shuttle operation. In the fused Autoware RViz view, red bounding boxes denote onboard detections and green bounding boxes denote infrastructure detections transmitted through the cloud. As highlighted by the golden rectangles, a pedestrian occluded from the vehicle by a light pole and signage is reliably detected by the elevated infrastructure sensors and integrated into the fused perception view. 
While this qualitative example illustrates the mechanism, we next provide a systematic quantitative evaluation of how infrastructure sensing improves observability across all safety-critical interactions in the roundabout.

\subsubsection{Ego-Centric Observability in Safety-Critical Roundabout Interactions}
\label{sec:v2i_observability}

We quantify whether infrastructure sensing improves the observability of safety-critical traffic participants around the autonomous shuttle in real roundabout conflict scenarios. The analysis uses the CoInfra-V2I sequence, where infrastructure-side 3D bounding boxes are annotated in the global frame and synchronized with the shuttle trajectory extracted from the onboard Autoware bags. The functional regions used in this study are shown in Fig.~\ref{Fig:CoverageRegion}.

\textbf{Analysis setup.}
We use the annotated 3D bounding boxes as reference objects and measure observability through LiDAR point support. For each object at each anchor timestamp, we count the number of LiDAR points inside the ground-truth box from (i) the onboard vehicle LiDAR only, (ii) the infrastructure nodes only, and (iii) the combined V2I point cloud, after removing ground points using the HD map.
We adopt class-dependent point-support thresholds to reflect different sensing requirements. VRUs are considered observable when at least \(10\) points fall inside the box, while vehicles require at least \(30\) points. This choice reflects the fact that vulnerable road users can often be functionally localized with fewer returns, whereas vehicles generally require denser support to provide reliable spatial extent for planning. The evaluation is ego-centric and conditioned on the shuttle state. When the shuttle is in the entry or ring region shown in Fig.~\ref{Fig:CoverageRegion}, we evaluate the most relevant counterpart actors in nearby conflict regions, including pedestrians in the closest crossing or waiting zones and vehicles in the conflicting entry or ring flow. To complement these structured conflict cases, we also evaluate all nearby VRUs and vehicles within \(30\,\mathrm{m}\) of the shuttle when it is located in either the entry or ring region.

\textbf{Metrics.}
We report two complementary metrics. \emph{Visibility rate} is the fraction of relevant counterpart objects that satisfy the class-dependent point-support threshold under vehicle-only, infrastructure-only, and combined V2I sensing. \emph{Recovery rate} measures the fraction of vehicle-only misses recovered by infrastructure sensing. We also report \emph{critical-frame completeness}, defined as the fraction of ego-centric conflict frames in which all relevant counterpart objects are simultaneously observable. This metric is particularly important for driving applications, since safe decision-making depends on complete awareness of all nearby conflict participants rather than isolated detections.

\begin{table*}[t]
\centering
\caption{Ego-centric observability under class-dependent point-support thresholds (VRU: \(N_p=10\), Vehicle: \(N_p=30\)). 
Object-level visibility is reported over all relevant counterpart objects, while critical-frame completeness indicates the fraction of frames in which all relevant counterpart objects are observable. 
Recovery denotes the fraction of onboard misses recovered by infrastructure sensing.}
\label{tab:v2i_ego_combined}
\scriptsize
\setlength{\tabcolsep}{4pt}
\begin{tabular}{l l c c c c c c c c c c}
\toprule
Scenario & Target & \(N_p\) & Obj. & Dist. & Veh. & Infra. & V2I & Recovery & Frames & Veh. comp. & V2I comp. \\
 &  &  & count & (m) & (\%) & (\%) & (\%) & (\%) &  & (\%) & (\%) \\
\midrule
AV entering \(\rightarrow\) VRU crossing                     & VRU     & 10 & 171  & 11.8 & 100.0 & 99.4 & 100.0 & ---   & 156 & 100.0 & 100.0 \\
AV entering \(\rightarrow\) VRU waiting                      & VRU     & 10 & 473  & 12.5 & 98.3  & 95.3 & 100.0 & 100.0 & 218 & 96.3  & 100.0 \\
AV entering \(\rightarrow\) vehicle in ring                  & Vehicle & 30 & 363  & 25.6 & 98.9  & 100.0 & 100.0 & 100.0 & 225 & 97.3  & 100.0 \\
AV in ring \(\rightarrow\) VRU crossing                      & VRU     & 10 & 292  & 13.2 & 80.5  & 92.8 & 99.7  & 98.2  & 236 & 75.8  & 99.6 \\
AV in ring \(\rightarrow\) VRU waiting                       & VRU     & 10 & 263  & 13.5 & 45.2  & 68.1 & 90.1  & 65.3  & 185 & 33.0  & 85.9 \\
AV in ring \(\rightarrow\) vehicle entering                  & Vehicle & 30 & 614  & 20.1 & 49.3  & 100.0 & 100.0 & 100.0 & 336 & 46.1  & 100.0 \\
AV in entry/ring \(\rightarrow\) nearby VRU within 30 m      & VRU     & 10 & 2426 & 17.4 & 66.0  & 70.9 & 89.9  & 57.1  & 1690 & 42.7  & 77.3 \\
AV in entry/ring \(\rightarrow\) nearby vehicle within 30 m  & Vehicle & 30 & 5100 & 19.7 & 63.6  & 99.6 & 99.6 & 98.9  & 2013 & 35.2  & 97.7 \\
\bottomrule
\end{tabular}
\end{table*}

\textbf{Results and discussion.}
Table~\ref{tab:v2i_ego_combined} shows that the benefit of V2I cooperation depends strongly on the interaction context. When the shuttle is approaching the roundabout, onboard sensing is already reliable in several cases, so infrastructure sensing mainly provides redundancy. In these entry scenarios, object-level visibility is already above \(98\%\) for waiting VRUs and vehicles in the ring, and remains \(100.0\%\) for crossing VRUs. Even so, combined V2I removes the remaining misses and raises critical-frame completeness to \(100.0\%\).

The benefit becomes decisive once the shuttle is already in the roundabout ring, where occlusion and line-of-sight competition are more severe. For \emph{AV in ring \(\rightarrow\) VRU crossing}, vehicle-only visibility is \(80.5\%\), while combined V2I raises it to \(99.7\%\); critical-frame completeness increases from \(75.8\%\) to \(99.6\%\). For \emph{AV in ring \(\rightarrow\) VRU waiting}, the gain is larger still: visibility rises from \(45.2\%\) to \(90.1\%\), and critical-frame completeness rises from \(33.0\%\) to \(85.9\%\). A similarly strong effect appears for \emph{AV in ring \(\rightarrow\) vehicle entering}, where visibility increases from \(49.3\%\) to \(100.0\%\) and critical-frame completeness from \(46.1\%\) to \(100.0\%\).

The two \(30\,\mathrm{m}\) nearby-object summaries show that the effect is broad rather than limited to a few hand-selected region pairs. When the shuttle is in the entry or ring region, combined V2I improves nearby-VRU visibility from \(66.0\%\) to \(89.9\%\) and nearby-vehicle visibility from \(63.6\%\) to \(99.6\%\). At the frame level, completeness rises from \(42.7\%\) to \(77.3\%\) for nearby VRUs and from \(35.2\%\) to \(97.7\%\) for nearby vehicles. These aggregate results are consistent with the structured conflict cases and show that the observability benefit is not confined to isolated examples.

This is the central empirical finding of the paper. Multi-node infrastructure perception, operating under the measured 5G latency regime, materially improves safety-relevant observability in exactly the conflict scenarios where vehicle-only sensing is most limited. The value of cooperative infrastructure perception should therefore be assessed not only through detector AP, where global fusion already dominates (Section~\ref{sec:perception_benchmarks}), but through its ability to restore complete situational awareness in real interaction contexts.

\subsection{Limitations}
\label{sec:limitations}
Several limitations should be considered when interpreting the results. First, the CoInfra-V2I subset is smaller than CoInfra-I2I and is currently limited to clear-weather shuttle runs. The observability study should therefore be viewed as a transportation-facing validation of the V2I pipeline rather than as a comprehensive benchmark across environmental conditions. Second, the ego-centric observability analysis uses LiDAR point-support thresholds as a practical proxy for situational awareness. This yields a compact and interpretable evaluation, but it does not replace full closed-loop assessment with downstream planning and control. Finally, the strongest global-fusion benchmarks do not exactly match the low-bandwidth object-level operating mode used by the deployed real-time system. However, the communication results suggest that point-cloud-based fusion may still be feasible within the measured 5G latency envelope, and GlobalLidar already performs close to GlobalLidarCam in many cases. Evaluating this intermediate operating point in a fully deployed real-time system remains future work.

\section{Conclusion}
\label{sec:conclusion}

This paper presented CoInfra, a deployable cooperative infrastructure perception platform, dataset, and open-source system stack for studying vehicle--infrastructure cooperation under realistic communication constraints and adverse-weather conditions. Beyond introducing a new benchmark, the paper argues that real-world cooperative infrastructure perception should be studied as an integrated sensing system, and that its practical value should be assessed not only through detector accuracy but also through safety-relevant observability in real traffic interactions.

The results support this argument along the full deployment pipeline. Real-world 5G measurements show that compact object-level communication is practical for multi-node operation, while richer geometric payloads such as cropped point clouds remain feasible at higher latency and raw image transmission does not scale. Trace-driven synchronization experiments show that adaptive fusion with lightweight state prediction suppresses tail latency, reducing P99 reaction time from 75.7~ms to 57.0~ms while preserving full fusion completeness. Cooperative perception benchmarks on CoInfra-I2I show that multi-node fusion substantially improves detection performance, with the largest gains concentrated on sparse and safety-critical objects, especially VRUs in conflict-relevant regions. These benchmark and communication results further suggest that the best deployment operating point may not be spatially uniform: compact late fusion is attractive as the default scalable mode, while richer point-cloud-based or feature-based fusion may be most valuable in selected high-risk regions.

Most importantly, the ego-centric observability analysis on CoInfra-V2I shows that the value of infrastructure cooperation extends beyond detector AP to the situational awareness that matters for safe driving. In structured roundabout conflict scenarios, combined V2I sensing raises critical-frame completeness from as low as 33\%--46\% with vehicle-only sensing to 86\%--100\%. This result shows that multi-node infrastructure perception can greatly restore awareness of safety-critical traffic participants precisely where onboard sensing is most limited by occlusion, range, and viewpoint constraints.

By releasing not only the dataset but also the full deployable stack (including hardware documentation, calibration and synchronization tools, fleet management software, and an Autoware-compatible ROS perception pipeline), we aim to lower the barrier to replicating, extending, and deploying cooperative infrastructure perception systems in the real world. Future work will expand the V2I subset to more diverse weather conditions and vehicle types, investigate practical intermediate operating points such as real-time point-cloud-based fusion, and extend the evaluation toward cooperative tracking, motion prediction, and planning-level closed-loop assessment.

\section*{Data and Code Availability}
The CoInfra dataset, hardware documentation, calibration and synchronization tools, web-based management platform, data recording utilities, and ROS-based perception stack (with Autoware-compatible message formats) are publicly available at \url{https://github.com/NingMingHao/CoInfra}.

\section*{Acknowledgments}
The authors gratefully acknowledge the financial support of the Natural Sciences and Engineering Research Council of Canada (NSERC) and MITACS, as well as the financial and technical support provided by Rogers Communications Inc.\ Canada.

\bibliographystyle{elsarticle-num}
\bibliography{bib}

\end{document}